\PassOptionsToPackage{dvipsnames}{xcolor}
\documentclass[11pt,letterpaper]{style}

\usepackage[numbers]{natbib}
\usepackage{graphicx}
\usepackage{booktabs}
\usepackage{amsmath,amsfonts,amssymb}
\usepackage{cleveref}
\usepackage{subcaption}
\usepackage{wrapfig}
\usepackage{multirow}
\usepackage{colortbl}
\usepackage{listings}
\usepackage{xparse}
\usepackage{fontawesome5}
\usepackage{float}
\usepackage{placeins}
\usepackage{threeparttable}

\graphicspath{{./}{fig/}{figures/}{plot/}{pdf/}{table/}}
\usepackage{amsthm}
\usepackage{tcolorbox}
\tcbuselibrary{listings, breakable}
\newtcblisting{mylisting}[1][]{
  listing only,
  breakable,
  colframe=gray!70,
  colback=gray!8,
  colbacktitle=gray!100,
  coltitle=white,
  fonttitle=\bfseries,
  boxrule=0.8pt,
  arc=2pt,
  title=#1,
  left=4pt,
  right=4pt,
  top=4pt,
  bottom=4pt,
  listing options={
    basicstyle=\ttfamily\small\color{black},
    keywordstyle=\color{black},
    stringstyle=\color{black},
    commentstyle=\color{black},
    breaklines=true,
    breakautoindent=false,
    breakindent=0em,
    breakatwhitespace=true,
    columns=fullflexible,
    numbers=none,
    frame=none,
    showstringspaces=false,
    keepspaces=true
  }
}
\usepackage{svg}
\tcbuselibrary{skins,breakable}
\tcbuselibrary{listingsutf8}
\usepackage{titletoc}

\usepackage{setspace}
\usepackage{pifont}
\usepackage{mathtools}
\usepackage{enumitem}
\usepackage{arydshln}
\usepackage{bbm}
\usepackage{lineno}
\usepackage{makecell}
\usepackage{adjustbox}
\usepackage[ruled,vlined]{algorithm2e}
\SetKwInput{KwRequire}{Require}
\SetKwInput{KwOutput}{Output}

\definecolor{mygray}{gray}{0.9}
\definecolor{syncol}{RGB}{243,246,249}
\definecolor{wildcol}{RGB}{215,240,235}
\definecolor{drop1}{RGB}{180,225,220}
\definecolor{drop2}{RGB}{150,210,200}
\definecolor{drop3}{RGB}{120,195,185}
\definecolor{drop4}{RGB}{95,180,170}
\definecolor{drop5}{RGB}{65,160,150}
\definecolor{lightblue}{RGB}{210,230,250}

\definecolor{myblue1}{HTML}{0171DC}
\definecolor{myblue2}{HTML}{013978}

\NewDocumentEnvironment{minted}{O{} m +b}{%
}{}

\renewcommand\Authfont{\centering\normalfont\bfseries\fontsize{11}{15}\selectfont}
\renewcommand\Affilfont{\centering\normalfont\fontsize{10}{15}\selectfont}

\newcommand{\pare}{\textbf{\textsc{Pare}}}
\newcommand{\colorpare}{\textsc{{\color{YellowGreen}P}roactive {\color{YellowGreen}A}gent {\color{YellowGreen}R}esearch {\color{YellowGreen}E}nvironment}}
\newcommand{\bench}{\textbf{\textsc{Pare-Bench}}}

\newcommand{\pear}[1]{\includegraphics[scale=#1]{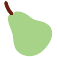}}%

\title{\pear{0.5}~\colorpare \\Simulating Active Users to Evaluate Proactive Assistants}
\runningtitle{\textbf{P}roactive \textbf{A}gent \textbf{R}esearch \textbf{E}nvironment}

\author{%
    {\Authfont
    \textbf{Deepak Nathani}\textsuperscript{1} \quad
    \textbf{Cheng Zhang}\textsuperscript{4} \quad
    \textbf{Chang Huan}\textsuperscript{1} \quad
    \textbf{Jiaming Shan}\textsuperscript{1} \quad
    \textbf{Yinfei Yang} \textsuperscript{2} \quad
    \textbf{Alkesh Patel}\textsuperscript{2} \quad
    \textbf{Zhe Gan}\textsuperscript{2} \quad
    \textbf{William Yang Wang}\textsuperscript{1} \quad
    \textbf{Michael Saxon}\textsuperscript{3} \quad
    \textbf{Xin Eric Wang}\textsuperscript{1} \quad
    }\\
    {\Affilfont
    \textsuperscript{1} University of California, Santa Barbara \quad
    \textsuperscript{2} Apple, USA \quad
    \textsuperscript{3} University of Washington \\
    \textsuperscript{4} Independent Researcher, USA

    }
}

\begin{document}
\begin{abstract}
Proactive agents that anticipate user needs and autonomously execute tasks hold great promise as digital assistants, yet the lack of realistic user simulation frameworks hinders their development.
Existing approaches model apps as flat tool-calling APIs, failing to capture the stateful and sequential nature of user interaction in digital environments and making realistic user simulation infeasible.
We introduce  Proactive Agent Research Environment (\pare), a framework for building and evaluating proactive agents in digital environments.
\pare\ models applications as \emph{finite state machines} with stateful navigation and state-dependent action space for the user simulator, enabling \emph{active user simulation}.
Building on this foundation, we present \bench, a benchmark of 143 diverse tasks spanning communication, productivity, scheduling, and lifestyle apps, designed to test context observation, goal inference, intervention timing, and multi-app orchestration.
\end{abstract}
\newcommand{\TitleLinks}{%
    \vspace{8pt}
    {\noindent\absfont\fontsize{11}{13}\selectfont
    \faGithub\ Project Page: \url{https://github.com/deepakn97/pare}\par}%
    \vspace{2pt}
    {\noindent\absfont\fontsize{11}{13}\selectfont
    \faEnvelope\ Correspondence: \href{mailto:dnathani@ucsb.edu}{dnathani@ucsb.edu}, \href{mailto:ericxwang@ucsb.edu}{ericxwang@ucsb.edu}\par}%
}

\maketitle

\section{Introduction}


Autonomous agents are at the frontier of language model (LM) applications research.
Capable of independently acting and completing complex tasks through tool-use  and multi-step reasoning, agent-based systems are succeeding mere instruction-following chatbots \citep{ouyang2022traininglanguagemodelsfollow}
across many application areas, including coding \citep{yang2024sweagentagentcomputerinterfacesenable}, web browsing \citep{zhou2024webarenarealisticwebenvironment}, computer use \citep{xieOSWorldBenchmarkingMultimodal2024} and virtual assistants \citep{trivedi2024appworldcontrollableworldapps}.
%
%

Current LM agents operate in a reactive paradigm: \textit{reactive agents} perform actions in response to an explicit user request \citep{barres$t^2$BenchEvaluatingConversational2025a}.
%
However, users are sometimes unaware of which actions need completing, and specifying needs to an assistant increases cognitive load.

\textit{Proactive agents} overcome these shortcomings by observing the actions a user takes and the information she recieves to infer her goals.
%
%
For example, a \textit{proactive assistant}, after observing a user's half-finished shopping list and a text message from their roommate saying they are out of soap, adds soap'' to the shopping list without the user's request; a reactive assistant would not do so until the user asks.


%
Recent work has begun exploring proactive assistants using LMs \citep{luProactiveAgentShifting2024,yangProAgentHarnessingOnDemand2025,yangContextAgentContextAwareProactive2025}, however, all prior approaches suffer from the same key limitation: reliance on \textit{passive, out-of-loop users for evaluation}.
We submit that \textbf{proactive agents may only be faithfully evaluated through interaction}.
Because user behavior is shaped by the behavior of an assistant, in the absence of real or simulated users interacting with the assistant, goal completion of the user-assisant system cannot be assessed.




\begin{figure}[t]
    \centering
    \includegraphics[width=0.96\textwidth]{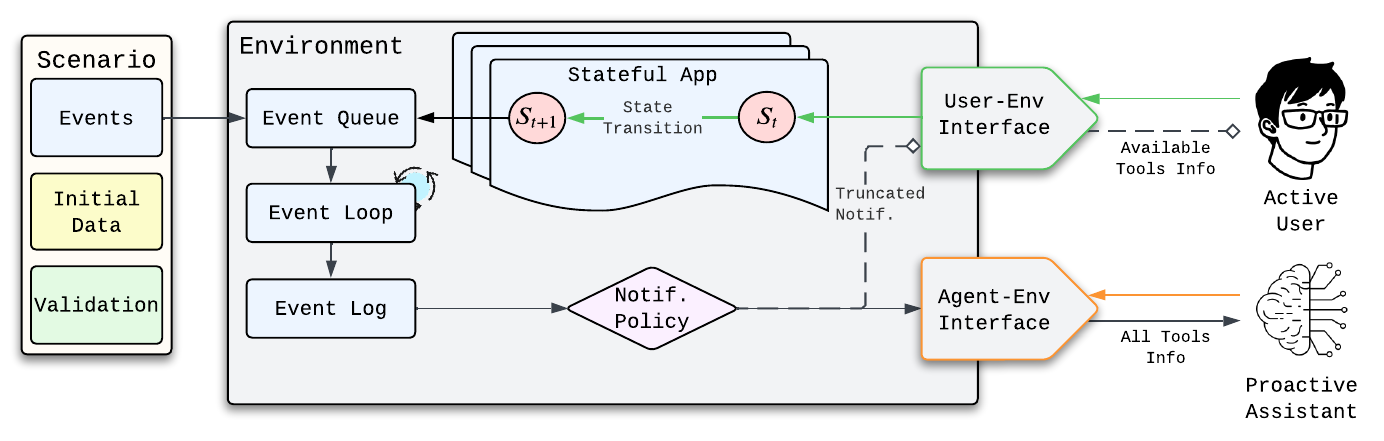}
    \caption{Overview of the \pare\ framework architecture. \pare\ framework consists of an event-based environment that models \textbf{Stateful App} transitions. The user-environment interface exposes selective tools based on the current state $S_t$ of the currently active app, and {\color{Green}user actions} result in app state transitions. Whereas the agent-environment interface exposes all tools as a flat API structure to enable efficient information gathering and task execution by Proactive Assistants.
    }
    \vspace{-1em}
    \label{fig:overview}
\end{figure}

To dynamically and scalably evaluate proactive assistants for goal-oriented impact, we introduce the \colorpare, \pare.
\pare~ extends the prior Agent Research Environment (ARE) \citep{froger2025arescalingagentenvironments} by creating \textbf{ecologically valid user simulator agents} to interact with proactive assistants under test.

Building these \textit{user simulator agents} requires deep modifications to the existing environment.
%
%
Unlike assistants which have unrestricted API access, real users must navigate the interfaces between and within an app to act and access information, as shown in \autoref{fig:tool_chain_comparison}.
%

To model this, the \textit{user simulator agents} in \pare\ interact with special
\textit{stateful app interfaces}.
User state is modeled by app-specific and global finite state machines (FSMs) which describe the interface logic real users would navigate.
Rather than being able to make any API call at a time, a user agent must navigate the FSM to select a function and populate its arguments (e.g. \autoref{fig:fsm_email}), thereby simulating the interface navigation process human users must take.
This asymmetry reflects real-world deployment, where a user is restricted to actions enabled by the current context on the screen, whereas a personal assistant on the same phone can directly access any functionality through backend APIs. An overview of the framework is presented in \autoref{fig:overview}.

We use \pare\ to build \bench, a benchmark of 143 diverse tasks designed to evaluate a proactive assistant's context observation, goal inference, intervention timing, and multi-app orchestration capabilities across communication, productivity, scheduling, and lifestyle
applications.
%
%
%
Using \bench\ we evaluate seven language model-based proactive assistants.


In summary, our contributions are as follows:
\begin{itemize}[topsep=0pt, noitemsep]
    \item \textbf{Proactive Agent Research Environment (\pare):} An asymmetric simulation framework where users navigate state-dependent actions while proactive agents access flat APIs. To the best of our knowledge, this is the first work to explore an asymmetric FSM-based environment design for evaluating proactive assistants.
    \item \textbf{Observe-Execute Proactive Architecture:} A two-phased agent design that separates the continuous monitoring from task execution, preserving the user's autonomy.
    \item \bench: A benchmark of 143 diverse tasks across four application domains, testing context observation, goal inference, intervention timing, and multi-app orchestration.
\end{itemize}

\section{Related Work}

\paragraph{LLMs as User Simulators} Generative Agents \citep{park2023generativeagentsinteractivesimulacra} demonstrated that LLMs can simulate passable human behavior in social environments.
This insight has been applied to agentic evaluation work, $\tau$-bench \citep{yao2024taubenchbenchmarktoolagentuserinteraction} uses LLM-simulated users to test tool-agent interaction in customer service domains, while ToolSandbox \citep{luToolSandboxStatefulConversational2025} introduces stateful tool execution with a built-in user simulator for conversational evaluation.
$\tau^2$-bench \citep{barres$t^2$BenchEvaluatingConversational2025a} extends this further by allowing simulated users to also take actions, creating a dual-control environment for technical support scenarios.

However, none of the above benchmarks target mobile environments or model the screen-by-screen navigation characteristics of phone interfaces.
\pare\ addresses this gap with FSM-based apps where simulated users navigate through state-dependent action spaces, mirroring real mobile interaction patterns.
Moreover, while these advances have produced capable reactive agents, they share a fundamental limitation that the user has to initiate the conversation, thus cannot be used as effective personal assistants.
We now turn our attention to \emph{proactive} agents, capable of observing user actions and a shared environment to infer tasks for the user and complete them proactively.

\paragraph{Mixed-Initiative Interfaces and Proactive Agents}

Researchers have long sought to study the problem of proactive personal assistants and mixed-initiative designs in the field of human-computer interaction.
\citet{maesAgentsReduceWork1994} proposed interface agents as ``personal apprentices`` that learn user preferences through observation and provide assistance with email, scheduling, and information filtering.
\citet{horvitzPrinciplesMixedinitiativeUser1999} formalized the principles for mixed-initiative interface design, emphasizing that agent interventions must be calibrated to confidence and the cost of errors, while \citet{shneidermanDirectManipulationVs1997} highlighted the tension between direct user control and intelligent agent autonomy.
\citet{Rhodes1997WearableRemembrance} introduces the Remembrance Agent, a ``just-int-time`` information retrieval system that proactively surfaces relavant information based on the user's current context.
These foundational works established the core challenges that persist today in the age of LLM-based proactive agents.

Recent works have started exploring the use of LLMs as proactive assistants.
ProactiveAgent \citep{luProactiveAgentShifting2024} proposes an LLM-driven gym where user activities and agent proposals are represented as natural language descriptions, but neither the user nor the agent execute any tool calls. Their generated events are textual and describe outcomes, but are not executable; as a result, the gym cannot track state.
ContextAgent \citep{yangContextAgentContextAwareProactive2025} builds a context-aware agent system for proactive assistance based on sensory data, but evaluates it on static dataset samples rather than an interactive simulation.
ProAgent \citep{yangProAgentHarnessingOnDemand2025} extends this with on-demand sensory context, yet similarly lacks dynamic environment interaction.
These approaches capture the high-level intent of proactive interaction, however, lack the grounded execution semantics that are necessary for realistic evaluation, i.e., the simulated users do not navigate through app screens, tools do not modify persistent state, and agents cannot observe the consequences of their actions in a dyanmic environment.

We argue that meaningful proactive agent evaluation requires an environment where users interact through state-dependent actions, as they would on a real-world UI based system and agents execute actions that modify shared state.
\pare\ bridges this gap by combining proactive goal inference with grounded execution in a feature-rich stateful mobile environment.
Simulated users navigate FSM-based apps through state-dependent actions while proactive agents access APIs from all apps at once, which reflects the assymtery of real-world deployment.
This design enabled evaluation of the complete proactive assistance loop, observing user actions, inferring goals, proposing interventions, and executing multi-step tasks.

\section{Proactive Agent Research Environment}

\pare\ is built on top of the Agent Research Environment (ARE) platform \citep{froger2025arescalingagentenvironments}, which provides abstractions for building reactive agent environments with apps, scenarios, and verification.
ARE agents receive instructions from a simulated user, but this user cannot act on the environment directly. ARE agents are reactive, and do not infer user goals or propose actions, only performing as instructed.
\textbf{Without simulated users to interact with, we cannot evaluate proactive assistants.}
%

\pare\ extends ARE chiefly by implementing \textit{active user simulator agents} inside the simulation environment.
While a crude user simulator could be implemented simply by prompting a second agent to pretend to be a user and take turns with the assistant, as shown in \autoref{fig:tool_chain_comparison}, this approach would fail to model the actual constraints users face as opposed to assistants.

\pare\ facilitates this simulation of active users by implementing \textit{stateful agent interfaces} which simulated users can use, as well as asymmetrical message passing and observation APIs to model how users and assistants interact.
%


\subsection{Formal Definition}
\label{sec:definition}

A \textit{proactive agent} $\mathbf{A}$ works collaboratively with user $\mathbf{U}$ within a shared environment state space $\mathcal{S}$ and action space $\mathcal{A}$.
Both the user and agent make \textit{observations} in observation space $\mathcal{O}$.
We formalize this as a Stackelberg POMDP \citep{brero2024stackelbergpomdpreinforcementlearning,oliehoek2016concise} (detailed in \autoref{sec:stackelberg_pomdp}).
The user has some set of goals $\mathcal{G}_\mathbf{U}$ which are not explicitly stated, and takes a series of actions $\mathcal{A}_\mathbf{U}$ toward them. The proactive agent attempts to infer user goals from their actions and the evolving state of the environment, and then \textit{proposes} plans $\mathbf{p}=\{a_i\in\mathcal{A}\}$ to the user.
If the user believes a plan will accomplish one of their goals, they \textit{accept} the plan and the agent performs its actions.

The agent and the user follow a Stackelberg turn-taking structure \citep{stackelberg2011market}, where at time step $t$ the user first acts, then the agent develops its plan, a function of its own prior actions and its observations of the user's prior actions and the current environment state $S_t$, and the past sequence of environment \textit{events} $E_{t}$.
\begin{equation}\label{eq:plan}
    \mathbf{p}_t = f_\theta(\mathcal{O}_{\mathbf{A}_{t-1}}, \mathcal{A}_{\mathbf{A}_{t-1}},\mathcal{A}_{\mathbf{U}_{t-1}},E_{t})
\end{equation}
At most time steps, no action is proposed, $\mathbf{p}_t=\emptyset$.
All proposed plans are part of the agent's \textit{plan set} $\mathcal{P}_\mathbf{A}$.
The binary function $\mathtt{ACCEPT}_\mathbf{U}(\mathbf{p}_i,\mathcal{G}_\mathbf{U})$ is true if the user believes plan $\mathbf{p}_i$ will help achieve one or more of their goals, in which case the plan's actions are executed and added to the \textit{agent action set} $\mathcal{A}_\mathbf{A}$.
%
%
\begin{equation}
    \mathtt{ACCEPT}_\mathbf{U}(\mathbf{p}_i) \implies \mathcal{A}_{\mathbf{A}_t} = \mathcal{A}_{\mathbf{A}_{t-1}} \cup \{a_i \in \mathbf{p}_t\}
\end{equation}


After both the user and assistant stop, if some goal $\mathbf{g}_i$ in $\mathcal{G}_\mathbf{U}$ is successfully fulfilled following the combined actions of the user and agent, we say the binary indicator function $\mathtt{SUCCEED}(\mathbf{g}_i,\mathcal{A}_\mathbf{U}\cup\mathcal{A}_\mathbf{A},\mathcal{S}_\textrm{final}) = 1$ is true.
The goal of a proactive agent is to jointly maximize the goal success rate $R_\textrm{Succeed}$ and the plan acceptance rate $R_\textrm{Accept}$ as follows:
\begin{equation}
    R_\textrm{Accept} = \frac{1}{|\mathcal{P}_\mathbf{A}|}\sum_{\mathbf{p}_i\in\mathcal{P}_\mathcal{A}}\mathtt{ACCEPT}_\mathbf{U}(\mathbf{p}_i,\mathcal{G}_\mathbf{U})
\end{equation}
\begin{equation}
    R_\textrm{Succeed} = \frac{1}{|\mathcal{G}_\mathbf{U}|}\sum_{\mathbf{g}_i\in\mathcal{G}_\mathbf{U}}\mathtt{SUCCEED}(\mathbf{g}_i,\mathcal{A}_\mathbf{U}\cup\mathcal{A}_\mathbf{A},\mathcal{S}_\textrm{final})
\end{equation}

\subsection{User and assistant action interfaces}
\label{sec:interfaces}

A \pare\ environment includes a \textit{user agent} and \textit{assistant agent} which interact with a shared set of \textit{apps} and recieve environment-level messages corresponding to `external' events. The user moves first, and then the agent and user alternate taking actions (or doing nothing) until a maximum number of turns is reached or a completion signal from the environment is provided.

The \textit{user agent} in \pare\ is a modified version of the ReAct agent \citep{yao2023reactsynergizingreasoningacting} provided by ARE, which has restricted API access to faithfully model a user.
In particular, the model only has access to a limited set of actions at any time, governed by its current \textit{user state} (\autoref{sec:transitions}), and it has limited \textit{observation} compared to the assistant (\autoref{sec:observations}).

In principle, the user and the assistant can perform identical operations on the underlying database (and thus the global environment state $\mathcal{S}$), but through different interfaces: the assistant makes direct API calls, while the user must traverse app states to populate arguments and invoke functions.
%

These user state navigation actions represent distinct actions in $\mathcal{A}$, which can be observed by the assistant.
A core \texttt{system} app exposes navigation tools, such as open, switch, and, close app to the user and utility functions such as \texttt{current\_time} to both the user and the assistant.

%
%
%

Finally, an \texttt{AgentUserInterface} app is required to provide an interface between the assistant and the user agents.
Through which the agent can access \texttt{send\_message\_to\_user} to propose a plan, and the user can respond with \texttt{accept\_proposal} and \texttt{reject\_proposal}.

\subsection{User state and transitions}
\label{sec:transitions}

We represent the \textit{user state space} as a finite state machine (FSM). Each application has its own FSM which exposes user-only tools as an interface between the user simulator and the underlying ARE APIs.
Each state represents a screen in the app, and transitions represent behaviors such as filling forms in fields, navigating screens, and submitting requests.

\begin{figure}[t]
    \centering
    \includegraphics[width=\columnwidth]{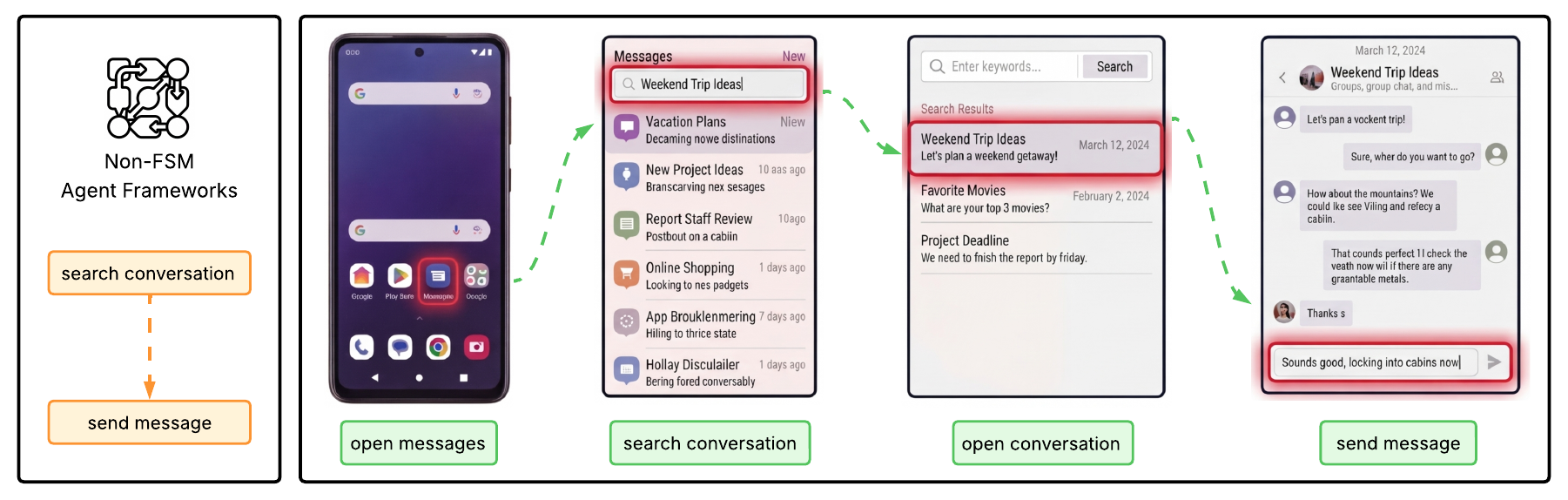}
    \caption{Comparison of tool call chains for sending a message. Non-FSM agent frameworks allow direct API calls (left), while \pare's FSM-based design requires the user to navigate through sequential app screens (right), matching how real users interact with mobile apps.}
    \label{fig:tool_chain_comparison}
\end{figure}

For example, \autoref{fig:tool_chain_comparison} shows how sending a message in a existing agent frameworks requires only two API calls (search conversation, send message), while in \pare\ the user must navigate through a realistic sequence of screens: opening the Messages app, searching for the conversation, opening it, and then sending the message.
We show the complete FSM state diagram for the Email app in \autoref{fig:fsm_email}.

%
%
%
%

In addition to app-level user state, \textit{environment-level user state}, i.e. which app is currently active, and which apps are running in the background, is also tracked.
The environment orchestrates app interactions and maintains the global navigation state.
%
%
%
The user has access to the stack of previously opened background apps and can return to them using app navigation commands from the environment-level \texttt{system} app.

Thus, \textit{all user actions are user FSM state transitions}.


\subsection{Environment events}
\label{sec:events}

Depending on the scenario, \pare\ environments contain \textit{events} which generate updates to the environment database as well as broadcast messages to the user and assistant.

%
Events can be scheduled at absolute timestamps or relative to other events, enabling dynamic scenarios where extenal such as a colleague replying to an email can be simulated, allowing scenarios spanning hours of simulated time to execute in minutes.
%

%

These events $E_t$ condition the agent and user's choice of actions at timestep $t$ as shown in \autoref{eq:plan}.

\subsection{User and assistant observation asymmetry}
\label{sec:observations}

The user and the assistant agent access different information to select their next action. \autoref{eq:plan} shows that the agent's choice of plan proposal (and actions) is a function of its previous observations $\mathcal{O}_{\mathbf{A}_{t-1}}$, the actions of itself and the user, $\mathcal{A}_{\mathbf{A}_{t-1}}$, $\mathcal{A}_{\mathbf{U}_{t-1}}$, and prior environment events $E_{t}$.
However, reflecting the limited observability that real users have into the internals of the system, the user agent only accesses its own past actions and observations, and cannot see the agent's actions:
\begin{equation}
    A_{\mathbf{U}_t} = f_\mathbf{U}(\mathcal{O}_{\mathbf{U}_{t-1}},\mathcal{A}_{\mathbf{U}_{t-1}},E_t)
\end{equation}



\pare\ routes \textit{event} notifications to the appropriate recipient in line with ARE's notification system.
For example, user actions are forwarded to the proactive agent as observations, agent messages are delivered to the user for acceptance or rejection, and environment events such as incoming emails or calendar reminders are broadcasted to both.

The level of verbosity at which event notifications are added to the event stack is configurable.
Users receive truncated notifications that mirror real mobile phone behavior, for instance, a new email notification displays the sender, subject line and truncated content.
In contrast, proactive agents receive the full serialized content.


%
%
%
%


\subsection{Scenarios}
\label{sec:scenarios}
Scenarios define an evaluation task for proactive agents.
Each scenario specifies an initial environment state, a sequence of initial events or triggers for proactive agents, and validation criteria for success.
Following ARE's scenario abstraction, \pare\ scenarios are divided into three components, app initialization, event flow construction, and oracle validation.

App initialization populates the environment with scenario-specific data such as contacts, calendar events, email threads etc., establishing the context.
The event flow defines scheduled environment events, and oracle events that specify expected proactive behavior.
%
To simulate real-world conditions, \pare\ inherits ARE's configurable distractor notifications and tool failure simulation, enabling scenarios with varying levels of difficulty.

\section{User Simulator and Proactive Assistant}
\label{sec:agents}

\pare\ orchestrates two LLM-based agents, and active user simulator and a \emph{proactive assistant} that interact through a Stackelberg turn-based loop \citep{stackelberg2011market,luToolSandboxStatefulConversational2025}, where the user acts first and the assistant observes and takes actions.
%
%
%

Both agents are built on ARE's BaseAgent \citep{froger2025arescalingagentenvironments} and follow the Reasoning $\rightarrow$ Action $\rightarrow$ Observation cycle introduced in ReAct \citep{yao2023reactsynergizingreasoningacting}.
\textbf{All agents in \pare\ can be configured to take multiple actions during their turn, and this is treated as a simulation parameter.}
In the following sections, we discuss the implementation details of the Active User Simulator and Proactive Assistant.

\subsection{User Simulator Agent}
As discussed in \autoref{sec:interfaces}, the user simulator has restricted access to the tools based on the current app state.
This constraint is modeled through the \emph{user-environment interface} and doesn't change the implementation of the ReAct agent.
%
%
%
%
At each turn, the agent receives a list of \emph{available apps}, \emph{current state}, and \emph{available tools} dynamically from the user-environment interface \autoref{fig:overview} based on the currently active app and navigation state.
%
%
The agent can explore the environment in an open-ended fashion or respond to incoming notifications, similar to \citet{park2023generativeagentsinteractivesimulacra}.

Beyond navigating apps, the user agent evaluates proposals from the proactive agent.
We model the \texttt{accept\_proposal} and \texttt{reject\_proposal} as context-dependent tools that are \textbf{only exposed to the user simulator when there is an assistant proposal awaiting review}.

%
The user simulator model is instructed to be strict about acceptance, rejecting vague proposals, misunderstanding the intent or context, or requiring capabilities beyond available apps.
%

\subsection{Proactive Assistant}
We implement the Proactive Assistant as a combination of two separate sub-agents: \emph{Observe} and \emph{Execute}.
This architecture design is motivated by the plan-then-execute pattern, which is common in agentic systems \citep{wu2023autogenenablingnextgenllm}.
%
%
%
The assistant system has three modes: \emph{observe}, \emph{awaiting confirmation}, and \emph{execute}, with \emph{awaiting confirmation} being a transient state where the assistant is waiting for the user to accept a proposal.

\paragraph{Observe Mode} In observe mode, the agent monitors user actions and environment notifications to identify proactive assistance opportunities.
The assistant only has access to read-only tools for information gathering, and two \emph{control actions}: \texttt{wait} to continue observing, or \texttt{send\_message\_to\_user} to propose assistance. Using either control action results in the termination of the assistant's turn.
%
%

%
If the assistant's proposal is accepted by the user, the assistant transitions to execution mode with the proposed goal as the initial task.
%
\paragraph{Execute Mode} The executor has access to the complete flat API across all apps included in the scenario, which enables it to complete tasks that span multiple applications.
The executor works autonomously, messaging the user only upon completion or if the task proves impossible.
After execution, the assistant returns to observe mode, ready to identify the next proactive assistance opportunity.
Similarly, if the user rejects a proposal, the agent returns to the observe mode without invoking the executor.

\section{\pare-Bench}


Building on the \pare\ framework, we create \bench, a benchmark of 143 scenarios designed to evaluate proactive agents across diverse contexts and apps.
%

While scenarios can be authored manually, this approach does not scale to the volume required for comprehensive evaluation.
Similar to recent works that employ LLMs to generate diverse tasks for LLM training and evaluation, we design an LLM-based scenario generation agent \citep{shen2024taskbenchbenchmarkinglargelanguage,xie2025agentsynthscalabletaskgeneration} that produces candidate scenarios.
Scenario generation can be broken down into four phases: story generation, initial app data population, building events flow, and validation.
Once our scenario generation agent has generated candidate scenarios, each scenario is verified by the authors of this work to ensure story coherence, correctness of the task execution validation criteria, and ensure the content of the scenario events is realistic.
We discuss the complete Scenario Generation Pipeline in \autoref{sec:scenario_generation}.
%
%

\bench\ scenarios use all the apps specified in \autoref{tab:apps}.
Since PASAgentUserInterface and System apps provide the core interface for assistant-user communication and inter-app navigation for the user simulator, all scenarios include these core applications.
The distribution of applications used across the benchmark is shown in \autoref{fig:app_usage}.
%

%
%
%
As discussed in \autoref{sec:scenarios}, we can increase the benchmark diversity by simulating tool failures and spurious notifications such as promotional or spam emails.
Fowllowing ARE \citep{froger2025arescalingagentenvironments}, noise events are scheduled using a Poisson process and interleaved with scenario events.
By varying the noise density, we can evaluate whether assistant models can distinguish signal from noise.

\section{Experiments}

We evaluate \textbf{seven LLMs} as proactive assistants on \bench: four closed-source models (Claude 4.5 Sonnet, GPT-5, Gemini 3 Pro, Gemini 3 Flash) and three open-weights models (Qwen 3 4B, Llama 3.2 3B, Gemma 3 4B).
We argue that proactive assistants should ideally be deployed as on-device models given the privacy implications of continuously observing user actions. We include frontier models to establish a performance ceiling and smaller models to evaluate the feasibility of on-device deployment.

\begin{table*}[ht]
    \centering
    \resizebox{\textwidth}{!}{%
    \begin{tabular}{lcccccc}
    \toprule
    \textbf{Model} & \textbf{Success@4 $\uparrow$} & \textbf{Success\textasciicircum 4 $\uparrow$} & \textbf{Success Rate $\uparrow$} & \textbf{Proposal Rate $\downarrow$} & \textbf{Acceptance Rate $\uparrow$} & \textbf{Read Actions} \\
    \midrule
    Claude 4.5 Sonnet        & 60.8\% & \textbf{18.2\%} & \textbf{42.0\% $\pm$ 1.0\%} & \textbf{12.8\% $\pm$ 0.4\%} & \textbf{78.2\% $\pm$ 0.8\%} & 20.2 $\pm$ 0.2 \\
    Gemini 3 Flash           & \textbf{64.3\%} & 16.1\% & 42.1\% $\pm$ 1.0\% & 19.1\% $\pm$ 0.4\% & 67.1\% $\pm$ 2.3\% & 21.7 $\pm$ 0.1 \\
    Gemini 3 Pro             & 59.4\% & 11.9\% & 35.1\% $\pm$ 0.9\% & 16.5\% $\pm$ 0.4\% & 71.2\% $\pm$ 1.6\% & 19.5 $\pm$ 0.2 \\
    GPT-5                    & 57.3\% & 17.5\% & 37.4\% $\pm$ 1.5\% & 28.1\% $\pm$ 0.3\% & 70.2\% $\pm$ 1.0\% & 20.6 $\pm$ 0.3 \\
    Qwen 3 4B Instruct       & 35.0\% & 6.3\%  & 18.5\% $\pm$ 1.3\% & 20.5\% $\pm$ 0.3\% & 63.7\% $\pm$ 1.6\% & 16.7 $\pm$ 0.1 \\
    Llama 3.2 3B Instruct    & 23.8\% & 1.4\%  & 10.0\% $\pm$ 0.4\% & 23.0\% $\pm$ 0.3\% & 58.4\% $\pm$ 0.3\% & 16.6 $\pm$ 0.3 \\
    Gemma 3 4B Instruct      & 7.7\%  & 0.7\%  & 3.0\% $\pm$ 0.4\%  & 14.2\% $\pm$ 0.3\% & 17.6\% $\pm$ 1.9\% & 8.8 $\pm$ 0.2 \\
    \bottomrule
    \end{tabular}
    }
    \caption{Model performance metrics on \bench\ across 4 runs. We report standard error ($\pm$). Proposal Rate follows lower is better, since too many proposals can annoy the user. The best performing model is highlighted in bold for each metric.}
    \label{tab:model_metrics}
\end{table*}

\paragraph{Metrics} As defined in \autoref{sec:definition}, we evaluate proactive assistants on Acceptance Rate ($R_\textrm{Accept}$) and Task Execution Success ($R_\textrm{Succeed}$).
Since proactive tasks can exhibit high variance across runs, we report success using three variants: \textbf{Success@$k$} checks if the assistant succeeds in \textbf{at least one} of $k$ runs, \textbf{Success\textasciicircum $k$} requires success across \textbf{all $k$ runs} and captures model reliability, and \textbf{Success Rate} calculates the average across all runs.
We also track \emph{Proposal Rate}, the proportion of turns in which the assistant proposed a task. A high proposal rate paired with low acceptance suggests the assistant is too eager and fails to gather sufficient context before intervening.
We report the average number of \emph{read-only actions} per scenario as a proxy for information gathering by the assistant.

\paragraph{Configuration} We use GPT-5-mini as the user simulation model.
The simulation runs for a maximum of 10 turns.
The user agent is configured to run 1 iteration per turn, while the proactive agent receives 5 iterations for observe mode and 10 iterations for execute mode.
We use a symmetric setup where the same LLM powers both the observe and execute agents.

\subsection{Results}
\autoref{tab:model_metrics} presents results across all seven models.
We find that Gemini 3 Flash and Claude 4.5 Sonnet perform comparably at the top, with success rates of 42.1\% and 42.0\% respectively.
Among the open-weights models, Qwen 3 4B Instruct leads across all metrics, while Gemma 3 4B Instruct struggles considerably at only 3.0\% success rate.

\paragraph{Consistency} The gap between Success@4 and Success\textasciicircum 4 reveals how consistent models are across repeated runs.
Claude drops from 60.8\% to 18.2\% (a 3.3$\times$ reduction), but Llama exhibits a 17.0$\times$ reduction from 23.8\% to 1.4\%.
This indicates that smaller models are not only less capable but far less consistent.
%

\begin{figure*}[ht]
    \centering
    \includegraphics[width=0.96\textwidth]{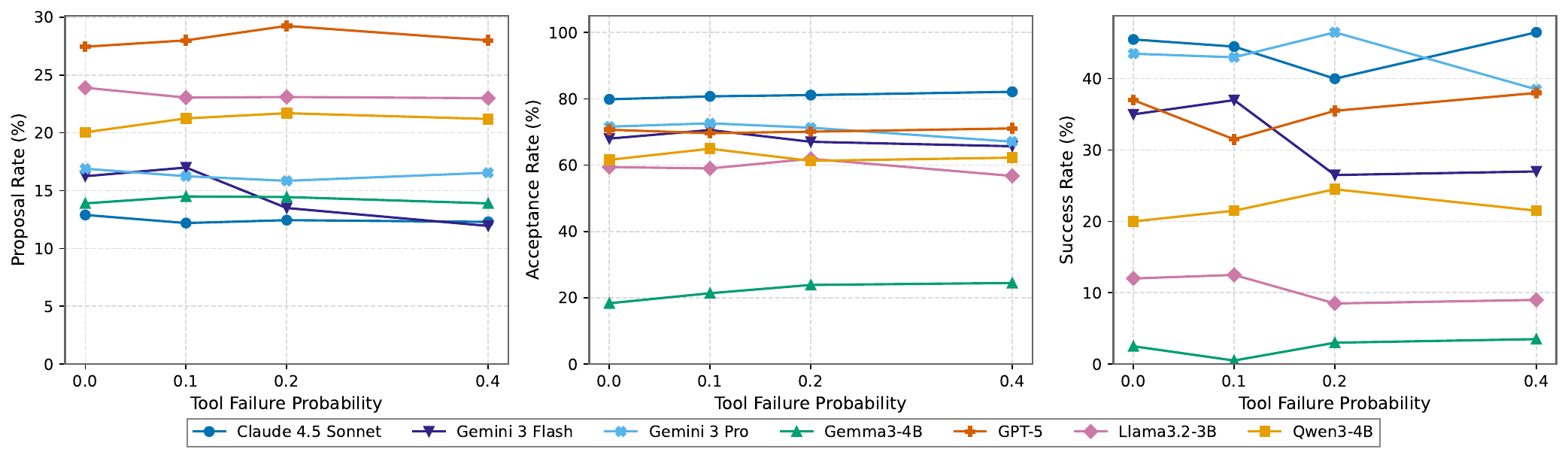}
    \caption{Proposal Rate, Acceptance rate, and Execution Success of different models with respect to tool failure probability evaluation across 4 runs.}
    \label{fig:robustness_tfp}
\end{figure*}

\begin{figure*}[ht]
    \centering
    \includegraphics[width=0.96\textwidth]{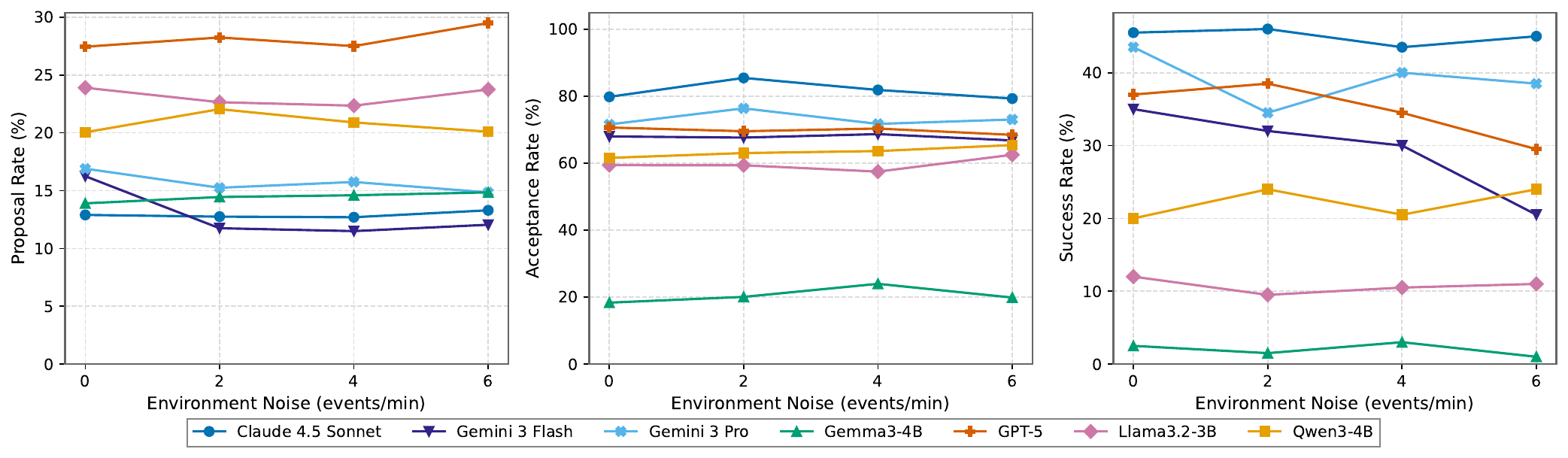}
    \caption{Proposal Rate, Acceptance rate, and Execution Success of different models with respect to environment event noise evaluation across 4 runs.}
    \label{fig:robustness_noise}
    \vspace{-0.2cm}
\end{figure*}

\paragraph{Proposal Quality and Efficiency} Acceptance rate ranges from 78.2\% for Claude to 17.6\% for Gemma.
Claude combines the lowest proposal rate (12.8\%) with the highest acceptance rate, indicating that it proposes only when it is confident.
Gemini 3 Flash achieves a comparable success rate to Claude (42.1\% vs 42.0\%), but with a higher proposal rate (19.1\%) and lower acceptance (67.1\%), which suggests that Gemini requires more proposals and more user interaction to achieve similar outcomes.
GPT-5 shows the highest proposal rate (28.1\%) paired with moderate acceptance (70.2\%), suggesting overconfidence in its goal inferences.
Gemma maintains the lowest proposal rate (14.2\%) despite its poor acceptance (17.6\%), which indicates that it is both passive and inaccurate when it does propose.
We present a finer-grained analysis of proposal decision patterns in \autoref{sec:ternary}.

\paragraph{Information Gathering and Execution} We observe a correlation between read actions and overall performance.
The top performing models, Gemini 3 Flash (21.7 actions), GPT-5 (20.6), and Claude (20.2), perform approximately 20\% more information gathering than Qwen and Llama (approximately 16.7), and nearly twice as many as Gemma (8.8).
This suggests that sufficient environmental observation is a prerequisite for accurate goal inference.
Interestingly, Qwen shows a relatively high acceptance rate (63.7\%) but a much lower success rate (18.5\%), which suggests that for smaller models, execution rather than goal inference is the primary bottleneck.

\subsection{Robustness under Environment Noise}

\paragraph{Tool Failure} We simulate tool failures with probability 0.1, 0.2, and 0.4 (\autoref{fig:robustness_tfp}).
Proposal and acceptance rates remain stable across all failure probabilities.
Success rates, however, reveal meaningful differences: Claude maintains 40-45\% even at 40\% failure probability, while Llama drops from 20\% to 8\% and Qwen remains stable at 18-20\%.

\paragraph{Environment Noise} We inject spurious notifications at 2, 4, and 6 events per minute to test whether models can distinguish relevant events from noise (\autoref{fig:robustness_noise}).
Proposal and acceptance rates remain stable across noise levels for all models.
We find that Claude maintains consistent success rates across all noise densities, while Gemini 3 Flash and GPT-5 degrade noticeably at higher noise.
Among smaller models, Qwen remains stable while Llama degrades, and Gemma stays near zero throughout.
This suggests that robustness to noisy environments varies considerably across models regardless of scale.

We further study the effect of user model choice on benchmark results in \autoref{sec:user_model_study}.

\section{Conclusion}
We introduce \pare, a framework for evaluating proactive assistants through active user simulation with FSM-based stateful app interfaces, \bench, a benchmark of 143 tasks testing goal inference, intervention timing, and multi-app orchestration, and the Observe-Execute agent architecture. Our experiments show that even the best frontier models achieve only 42\% success rate, and that for smaller models execution is the primary bottleneck, which the Observe-Execute design can partially address. We discuss limitations and future work in \autoref{sec:limitations}.

\section*{Impact Statement}

This work introduces \pare, a framework and benchmark for proactive AI assistants, advancing research in goal inference, intervention timing, and multi-app orchestration.
We strongly believe that since Proactive agents require observing user actions, they should use models deployed on edge devices which eliminates the need for uploading user data to external servers.
Moreover, we make a conscious decision of modeling user actions as APIs.
This API-level abstraction provides a natural privacy boundary—agents observe what actions occur, not everything visible on screen.

With advent of Proactive Assistants, another concern is the loss of autonomy of the human users.
We believe proactive assistants can be valuable if developed with human as the sole point of control.
This sets up the motivation for our Observe-then-Execute architecture which forces the agent to ask permission before executing any task.

We release \pare\ framework and \bench\ to enable reproducible research and encourage the community to build capable yet privacy-preserving proactive assistants.

\bibliographystyle{unsrtnat}
\bibliography{reference}

\appendix
\section{Limitations and Future Work}
\label{sec:limitations}

\paragraph{Limitations}
\pare\ models app interactions as tool calls rather than visual screen interactions.
While this is a deliberate design choice that provides a natural privacy boundary and enables scalable evaluation, it means we do not test the visual grounding capabilities that would be required in a real multimodal deployment.

Our user simulation is LM-based, and while our user model study (\autoref{sec:user_model_study}) shows that relative rankings of proactive assistants are preserved across different user models, simulated users may not capture all nuances of real human behavior such as fatigue, emotional state, and multitasking patterns.
Moreover, our user simulation does not model individual differences in user personality and trust levels.
In practice, some users are more trusting and would accept proposals without verifying the underlying content, while others actively reject proposals if the intervention timing does not match their preferences.
Modeling this personalization through theory-of-mind approaches \citep{zhou2026tomsweusermentalmodeling} remains an open challenge.

In our main experiments, we use a symmetric setup where the same model powers both the observe and execute agents.
The intended deployment pattern is asymmetric, with a smaller on-device model for continuous observation and a larger model invoked for execution, but we do not evaluate this configuration.

While \bench\ includes 143 scenarios spanning multiple apps and domains, the benchmark may not cover all real-world proactive assistance situations, such as long-horizon tasks spanning multiple days or tasks requiring world knowledge beyond the app ecosystem.

\paragraph{Future Work}
A natural next step is to evaluate asymmetric observe-execute configurations.
Our finding that execution is the primary bottleneck for smaller models motivates pairing a small on-device observation model with a larger execution model.
For example, a quantized 4B parameter model could run continuously on-device for observation, while a cloud-hosted model is invoked only when the user accepts a proposal, keeping user data on-device until explicit consent is given.

Expanding the app ecosystem with more complex apps and richer inter-app dependencies would enable more realistic multi-app orchestration scenarios.
For instance, adding a banking app with transaction capabilities would introduce safety-critical actions where the cost of incorrect proactive intervention is significantly higher, testing whether agents can calibrate their proposal confidence accordingly.

The proactive assistance task can be naturally modeled as a POMDP, where the hidden state includes the user's latent goals and the agent must select actions based on partial observations.
The acceptance and success metrics defined in \autoref{sec:definition} can serve as reward signals for training proactive agents using reinforcement learning, potentially closing the gap between frontier and smaller models.

Incorporating UI screenshots alongside API-based interactions would enable evaluation of multimodal proactive agents.
This would bridge the gap between our API-level abstraction and real-world deployment, where agents must process visual context from the user's screen to infer goals and timing for intervention.

\section{Proposal Decision Analysis}
\label{sec:ternary}

The binary accept/reject analysis in the main experiments does not capture the full picture of user-agent interaction.
We analyze the trajectories generated by the main experiment run (GPT-5-mini as user model, 4 runs per scenario) and categorize user responses into three decisions: \texttt{accept}, \texttt{reject}, and \texttt{gather context}.
In the \texttt{gather context} case, the user ignores the proposal and continues exploring the environment to gather more information on their own before deciding.
For example, when the user receives a truncated notification about a new email, the proactive assistant may propose to schedule a meeting based on the email.
The user, having only seen the truncated notification, may choose to ignore the proposal and navigate to the email app to read the full content before deciding.
If instead the assistant had waited for the user to read the email first, the user would have had full context and the proposal would have a higher chance of direct acceptance.
A high \texttt{gather context} rate indicates that the assistant's intervention timing is premature.
We also track \texttt{truncated} proposals, where the simulation ended due to the maximum turn limit before the user could resolve the proposal.
Results are shown in \autoref{tab:ternary_decisions}.

\begin{table}[ht]
\centering
{%
\begin{tabular}{lcccc}
\toprule
\textbf{Model} & \textbf{Accept $\uparrow$} & \textbf{Reject} & \textbf{Gather Context $\downarrow$} & \textbf{Truncated} \\
\midrule
Claude 4.5 Sonnet     & \textbf{72.1\% $\pm$ 1.4\%} & 7.8\% $\pm$ 0.8\% & \textbf{17.8\% $\pm$ 1.9\%} & 2.3\% $\pm$ 0.7\% \\
Gemini 3 Flash        & 61.5\% $\pm$ 1.9\% & 14.7\% $\pm$ 1.9\% & 18.3\% $\pm$ 0.8\% & 5.5\% $\pm$ 0.9\% \\
Gemini 3 Pro          & 65.6\% $\pm$ 1.5\% & 13.3\% $\pm$ 1.0\% & 15.1\% $\pm$ 1.0\% & 5.9\% $\pm$ 1.0\% \\
GPT-5                 & 64.1\% $\pm$ 1.3\% & 7.4\% $\pm$ 0.3\%  & 23.4\% $\pm$ 1.4\% & 5.1\% $\pm$ 0.4\% \\
Qwen 3 4B Instruct    & 56.3\% $\pm$ 1.6\% & 14.1\% $\pm$ 1.0\% & 26.5\% $\pm$ 1.1\% & 3.2\% $\pm$ 0.7\% \\
Llama 3.2 3B Instruct & 49.9\% $\pm$ 0.7\% & 15.3\% $\pm$ 1.0\% & 29.1\% $\pm$ 0.3\% & 5.8\% $\pm$ 0.3\% \\
Gemma 3 4B Instruct   & 16.0\% $\pm$ 1.5\% & 6.4\% $\pm$ 1.5\%  & 74.7\% $\pm$ 2.6\% & 2.8\% $\pm$ 0.2\% \\
\bottomrule
\end{tabular}
}
\caption{Ternary decision distribution across models. We report standard error ($\pm$) across 4 runs. The best performing model is highlighted in bold for each metric.}
\label{tab:ternary_decisions}
\end{table}

\paragraph{Decision Distribution}
Claude has the highest direct accept rate (72.1\% $\pm$ 1.4\%) and the lowest reject rate (7.8\% $\pm$ 0.8\%), indicating that its proposals closely match user simulator intent.
Gemma stands out with 74.7\% $\pm$ 2.6\% of its proposals triggering \texttt{gather context}.
This suggests that the vast majority of Gemma's interventions are premature, and only 16.0\% of its proposals are directly accepted.
Among frontier models, reject rates split into two groups: Claude and GPT-5 at approximately 7-8\%, versus Gemini Flash and Pro at 13-15\%.
Smaller models such as Qwen (26.5\%) and Llama (29.1\%) trigger \texttt{gather context} more often than frontier models, suggesting their proposals are often premature.

\begin{table}[ht]
\centering
{%
\begin{tabular}{lccc}
\toprule
\textbf{Model} & \textbf{G $\rightarrow$ Accept $\uparrow$} & \textbf{G $\rightarrow$ Reject $\downarrow$} & \textbf{G $\rightarrow$ Truncated} \\
\midrule
Claude 4.5 Sonnet     & \textbf{29.8\% $\pm$ 4.3\%} & \textbf{4.6\% $\pm$ 3.0\%} & 65.6\% $\pm$ 4.4\% \\
Gemini 3 Flash        & 25.5\% $\pm$ 1.2\% & 9.5\% $\pm$ 1.4\%  & 65.0\% $\pm$ 1.2\% \\
Gemini 3 Pro          & 28.7\% $\pm$ 3.0\% & 5.6\% $\pm$ 2.4\%  & 65.7\% $\pm$ 0.7\% \\
GPT-5                 & 25.2\% $\pm$ 1.5\% & 3.7\% $\pm$ 1.1\%  & 71.1\% $\pm$ 1.4\% \\
Qwen 3 4B Instruct    & 25.5\% $\pm$ 2.4\% & 9.7\% $\pm$ 2.8\%  & 64.8\% $\pm$ 0.9\% \\
Llama 3.2 3B Instruct & 27.4\% $\pm$ 1.4\% & 10.2\% $\pm$ 0.5\% & 62.4\% $\pm$ 1.9\% \\
Gemma 3 4B Instruct   & 10.1\% $\pm$ 1.0\% & 10.4\% $\pm$ 0.8\% & 79.5\% $\pm$ 0.9\% \\
\bottomrule
\end{tabular}
}
\caption{Gather context resolution. We report the percentage of proposals that entered the information gathering phase and how they eventually resolved. Standard error ($\pm$) across 4 runs.}
\label{tab:gather_context}
\end{table}

\paragraph{Gather Context Resolution}
\autoref{tab:gather_context} shows what percentage of proposals that entered the information gathering phase eventually resolved into \texttt{accept}, \texttt{reject}, or were \texttt{truncated}.
Across all models, the dominant outcome is truncation (62-80\%), meaning the user runs out of turns while still gathering information.
This is a consequence of the 10-turn limit in our experimental configuration; this pattern would be interesting to study with higher turn budgets.
The conversion from \texttt{gather context} to \texttt{accept} is consistent across frontier models (25-30\%) but drops to 10.1\% for Gemma, indicating that even after gathering additional context, Gemma's proposals rarely become acceptable.
The conversion to \texttt{reject} is low for Claude and GPT-5 (3.7-4.6\%) but higher for smaller models (9-10\%), indicating that after gathering more context the user more often finds flaws in weaker models' proposals.

\section{User Simulator Ablation}
\label{sec:user_model_study}

\begin{table*}[ht]
  \centering
  \resizebox{\textwidth}{!}{%
  \begin{tabular}{lcccccc}
  \toprule
  \textbf{Proactive Model} & \textbf{Success@4 $\uparrow$} & \textbf{Success\textasciicircum 4 $\uparrow$} & \textbf{Success Rate $\uparrow$} & \textbf{Proposal Rate $\downarrow$} & \textbf{Acceptance Rate $\uparrow$} & \textbf{Read Actions} \\
  \midrule
  \rowcolor{syncol} \multicolumn{7}{c}{\textbf{User Model: GPT-5-mini}} \\
  Claude 4.5 Sonnet   & \textbf{64.0\%} & 22.0\% & \textbf{45.5\% $\pm$ 2.2\%} & \textbf{12.9\% $\pm$ 0.9\%} & \textbf{79.8\% $\pm$ 2.4\%} & 20.3 $\pm$ 0.0 \\
  GPT-5               & 54.0\% & \textbf{24.0\%} & 37.0\% $\pm$ 2.4\% & 27.5\% $\pm$ 0.3\% & 70.7\% $\pm$ 1.4\% & 20.5 $\pm$ 0.3 \\
  Gemini 3 Flash      & 56.0\% & 10.0\% & 35.0\% $\pm$ 2.4\% & 16.3\% $\pm$ 0.5\% & 68.0\% $\pm$ 4.5\% & 19.2 $\pm$ 0.4 \\
  Qwen 3 4B Instruct  & 36.0\% & 8.0\%  & 20.0\% $\pm$ 1.4\% & 20.1\% $\pm$ 0.3\% & 61.6\% $\pm$ 2.5\% & 16.7 $\pm$ 0.2 \\
  \midrule
  \rowcolor{syncol} \multicolumn{7}{c}{\textbf{User Model: Claude 4.5 Sonnet}} \\
  Claude 4.5 Sonnet   & 40.0\% & \textbf{14.0\%} & \textbf{26.0\% $\pm$ 2.4\%} & \textbf{10.6\% $\pm$ 0.2\%} & \textbf{49.5\% $\pm$ 1.1\%} & 17.0 $\pm$ 0.2 \\
  GPT-5               & \textbf{42.0\%} & 10.0\% & 25.0\% $\pm$ 2.9\% & 21.5\% $\pm$ 0.8\% & 34.7\% $\pm$ 1.2\% & 18.3 $\pm$ 0.5 \\
  Gemini 3 Flash      & 38.0\% & \textbf{14.0\%} & 26.0\% $\pm$ 1.8\% & 14.6\% $\pm$ 0.2\% & 32.9\% $\pm$ 2.1\% & 18.3 $\pm$ 0.3 \\
  Qwen 3 4B Instruct  & 28.0\% & 4.0\%  & 13.5\% $\pm$ 2.1\% & 16.8\% $\pm$ 0.3\% & 23.9\% $\pm$ 0.8\% & 15.6 $\pm$ 0.1 \\
  \midrule
  \rowcolor{syncol} \multicolumn{7}{c}{\textbf{User Model: Qwen 3 4B}} \\
  Claude 4.5 Sonnet   & 52.0\% & \textbf{22.0\%} & 36.5\% $\pm$ 1.0\% & \textbf{11.0\% $\pm$ 0.2\%} & \textbf{80.0\% $\pm$ 2.9\%} & 23.6 $\pm$ 0.1 \\
  GPT-5               & 64.0\% & 16.0\% & 36.5\% $\pm$ 2.6\% & 21.7\% $\pm$ 1.5\% & 77.0\% $\pm$ 1.8\% & 24.1 $\pm$ 0.3 \\
  Gemini 3 Flash      & \textbf{66.0\%} & 16.0\% & \textbf{44.5\% $\pm$ 2.6\%} & 14.3\% $\pm$ 0.8\% & 74.6\% $\pm$ 3.5\% & 25.8 $\pm$ 0.5 \\
  Qwen 3 4B Instruct  & 28.0\% & 0.0\%  & 11.5\% $\pm$ 1.0\% & \textbf{9.6\% $\pm$ 0.3\%}  & 76.4\% $\pm$ 4.2\% & 20.9 $\pm$ 0.3 \\
  \bottomrule
\end{tabular}
}
\caption{User model study on the ablation split (50 scenarios, 4 runs each). We vary the user simulation model while keeping the proactive assistant models fixed. Each section groups results by user model. The best performing proactive model is highlighted in bold for each metric within each group.}
\label{tab:user_model_study}
\end{table*}

To understand how user model choice affects benchmark results, we vary the user simulation model while keeping the proactive assistant models fixed. We compare three user models: GPT-5-mini (our default from the main experiments), Claude 4.5 Sonnet, and Qwen 3 4B. Each user model is paired with four proactive assistants (Claude 4.5 Sonnet, GPT-5, Gemini 3 Flash, and Qwen 3 4B) on the ablation split (50 scenarios, 4 runs each). Results are shown in \autoref{tab:user_model_study}.

\paragraph{User Model Sensitivity} Since the proactive agent observes and reacts to user actions, the user model can indirectly affect the proactive assistant's scenario success rates. We find that Claude as the user is the strictest evaluator: acceptance rates drop to the 23-49\% range and success rates compress to 13-26\% across all proactive models. Qwen as the user is the most permissive, with acceptance rates uniformly high (74-80\%) across all proactive models, yet success rates still vary widely (11-44\%). GPT-5-mini sits in between, with sufficient spread in both acceptance and success rates to discriminate between models. This validates our choice of GPT-5-mini as the default user model for the main experiments.

\paragraph{Ranking Stability} Despite large shifts in absolute numbers, we find that the relative ordering of proactive assistants is preserved across all three user models. We observe no same-model bias: Claude achieves the highest acceptance rate regardless of which model powers the user (49.5\% with Claude user, 79.8\% with GPT-5-mini, 80.0\% with Qwen). Proposal rate patterns are also largely intrinsic to each proactive model, with Claude consistently proposing the least and GPT-5 the most, though absolute values shift slightly with different user models.

\paragraph{Observation Behavior} We observe that read actions increase significantly with Qwen as the user (20-26 range) compared to other user models (15-20). This suggests that user behavior patterns influence how much the proactive agent explores the environment. Even with high read actions and the most permissive user, Qwen as proactive assistant achieves 0\% Pass\textasciicircum 4, reinforcing that information gathering alone does not compensate for poor execution capability.

\section{Stackelberg POMDP Formulation}
\label{sec:stackelberg_pomdp}

We formalize \pare\ as a Stackelberg POMDP \citep{brero2024stackelbergpomdpreinforcementlearning,oliehoek2016concise} defined by the tuple $\mathcal{M} = \langle \mathcal{N}, \mathcal{S}, \{\mathcal{A}_i\}_{i \in \mathcal{N}}, T, R, \{\mathcal{O}_i\}_{i \in \mathcal{N}}, \mathcal{I} \rangle$, where $\mathcal{N} = \{\mathbf{U}, \mathbf{A}\}$ denotes the user (leader) and the proactive agent (follower).
The user drives the interaction by taking actions first, and the proactive agent observes user behavior before deciding whether to intervene.
Each component of the tuple is detailed below with examples drawn from the mobile environment.

\paragraph{State space ($\mathcal{S}$)}
The global state is decomposed as $\mathcal{S} = \mathcal{S}_{\text{app}} \times \mathcal{S}_{\text{global}} \times \mathcal{S}_{\text{db}} \times \mathcal{S}_{\text{history}}$.
$\mathcal{S}_{\text{app}}$ represents the current screen within the active app; for example, in the Contacts app this could be the contacts list, a specific contact's detail view, or the edit screen for that contact.
$\mathcal{S}_{\text{global}}$ tracks which app is currently in the foreground.
$\mathcal{S}_{\text{db}}$ captures the persistent data across all apps, such as contacts, emails, calendar events, and shopping cart items.
$\mathcal{S}_{\text{history}}$ maintains a bounded view of past events for each agent, including the navigation stack, recent actions, and notifications.

\paragraph{Action spaces ($\mathcal{A}_i$)}
The user's available actions are state-dependent: $\mathcal{A}_\mathbf{U}: \mathcal{S}_{\text{app}} \times \mathcal{S}_{\text{global}} \rightarrow 2^{\mathcal{A}_\mathbf{U}^{\text{all}}}$.
For example, from the contacts list the user can search contacts, open a contact, or create a new one, but cannot edit a contact without first navigating to the detail view.
When a proposal from the agent is pending, the user's action space is augmented with \texttt{accept\_proposal} and \texttt{reject\_proposal}.
The proactive agent's action space is state-independent: $\mathcal{A}_\mathbf{A} = \mathcal{A}_{\text{read}} \cup \mathcal{A}_{\text{propose}} \cup \{\texttt{wait}\}$, where $\mathcal{A}_{\text{read}}$ includes read-only queries across all apps, $\mathcal{A}_{\text{propose}}$ allows the agent to propose a task to the user, and \texttt{wait} continues observation without intervention.

\begin{figure}[h]
    \centering
    \includegraphics[width=0.48\textwidth]{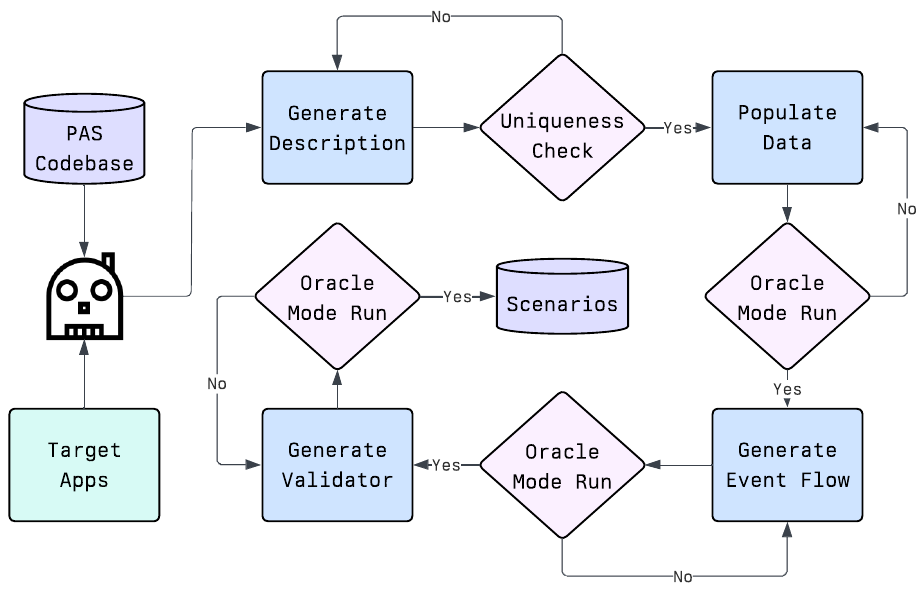}
    \caption{Scenario generation agent pipeline. Given \pare\ codebase and target apps, the agent generates scenarios through four stages: description, data population, event flow, and validation. Each stage (except description) runs in oracle mode for execution-based validation, with failed attempts looping back for retry. The uniqueness check ensures no duplicate scenarios.}
    \label{fig:scenario_generator}
\end{figure}
\paragraph{Observation spaces ($\mathcal{O}_i$)}
The user's observation $\mathcal{O}_\mathbf{U}(s)$ is a function of the current state only, and includes the current screen, available tools, any pending agent proposal, and truncated environment notifications.
For example, a new email notification shows only the sender and subject line, not the full content.
The agent's observation $\mathcal{O}_\mathbf{A}(s, a_\mathbf{U})$ is a function of both the state and the user's action, reflecting the Stackelberg structure.
The agent observes the user's executed actions, receives full environment notifications, and sees the user's response to any pending proposal.

\paragraph{Transition function ($T$)}
$T: \mathcal{S} \times \mathcal{A}_\mathbf{U} \times \mathcal{A}_\mathbf{A} \rightarrow \mathcal{S}$ defines the state dynamics.
The base model is deterministic: for example, the user calling \texttt{open\_contact("C001")} from the contacts list transitions $\mathcal{S}_{\text{app}}$ from \texttt{ContactsList} to \texttt{ContactDetail("C001")}.
A stochastic extension supports configurable tool failure probability for robustness experiments (\autoref{fig:robustness_tfp}).

\paragraph{Reward function ($R$)}
\pare\ uses a dual reward structure: $R = (R_\textrm{Succeed}, R_\textrm{Accept})$, corresponding to the metrics defined in \autoref{sec:definition}.
$R_\textrm{Succeed}$ is a terminal binary reward that equals 1 if the user's goals $\mathcal{G}_\mathbf{U}$ are fulfilled in the final environment state, as verified by the scenario oracle.
$R_\textrm{Accept}$ is a per-step reward: $+1$ for accepted proposals, $-1$ for rejected proposals, and $0$ otherwise.

\paragraph{Instruction space ($\mathcal{I}$)}
$\mathcal{I} = \mathcal{I}_\mathbf{U} \times \mathcal{I}_\mathbf{A}$ specifies task instructions for each agent at the start of an episode.
The user receives a natural language goal description, for example ``You received an email about a meeting with Alice. Schedule it on your calendar.''
The agent receives generic proactive assistance instructions that describe its role as an observer and helper.

\paragraph{Episode dynamics}
Each turn follows the Stackelberg structure: the user acts first, the agent observes the user's action, then the agent acts.
The environment updates the state and computes rewards after both agents have acted.
Episodes terminate when the maximum number of turns $T_{\max}$ is reached or the environment signals completion.

\section{Scenario Generation Pipeline}
\label{sec:scenario_generation}

\subsection{Scenario Generation Agent}
As discussed in \autoref{sec:scenarios}, a scenario in \pare\ consists of three components: initial app states, environment event flow and validation.

Following this abstraction, we design our scenario generation agent in a four-stage pipeline (\autoref{fig:scenario_generator}), scenario description generation, initial app state population, building events flow and validation function.
We use a code generation agent based on Claude-Agent-SDK\footnote{https://github.com/anthropics/claude-agent-sdk-python}, which receives the \pare\ codebase in a read-only context and a selection of target apps that it has to include in the scenario.

In the first stage, the agent generates a natural language description of the scenario, which is checked against existing scenario description for uniqueness using LLM-as-a-judge \citep{gu2025surveyllmasajudge, zheng2023judgingllmasajudgemtbenchchatbot}.

\begin{wrapfigure}{r}{0.48\columnwidth}
    \centering
    \includegraphics[width=0.48\columnwidth]{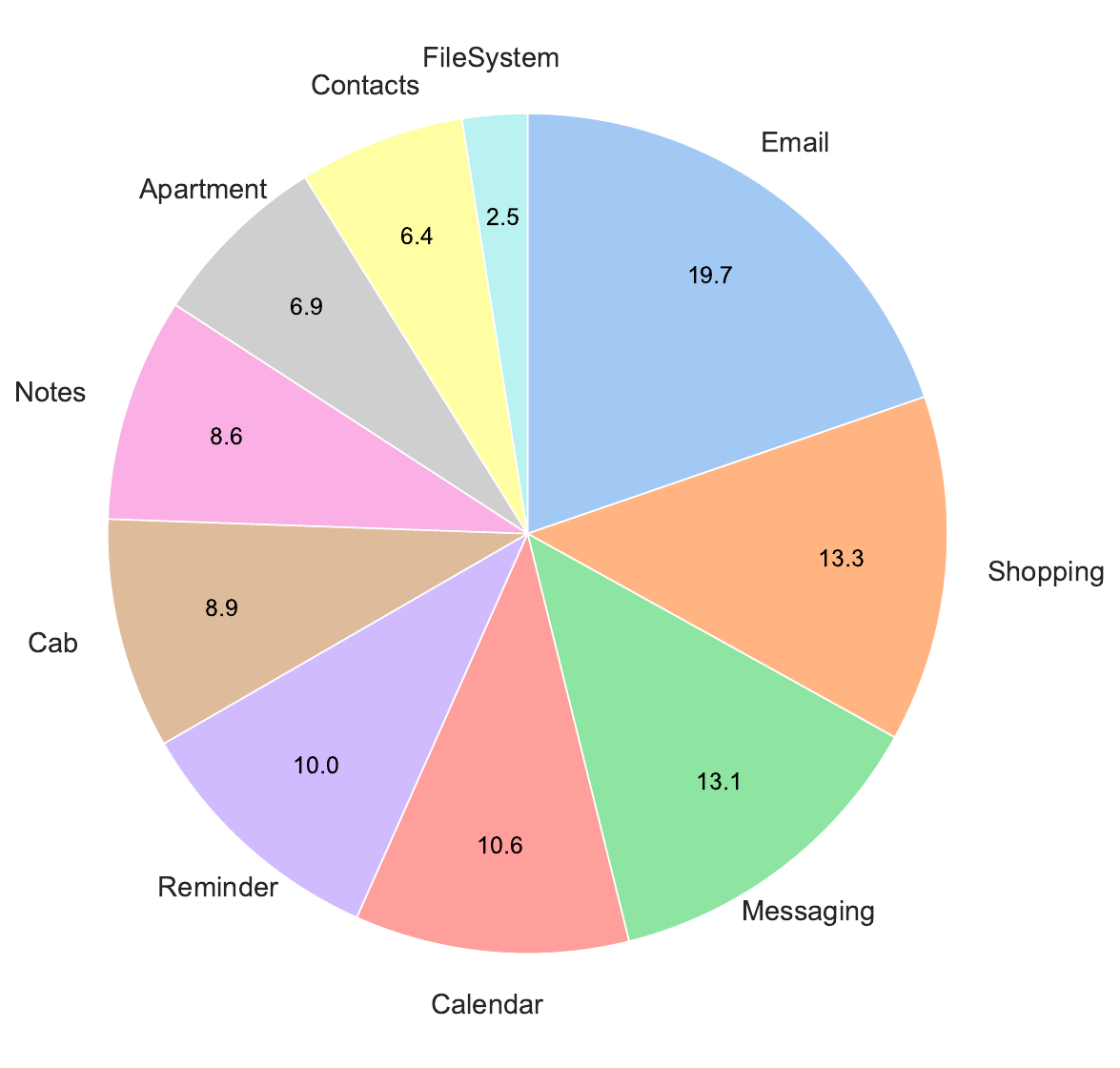}
    \caption{App usage statistics across 143 scenarios in \bench. PASAgentUserInterface and System apps are excluded since they are included in all scenarios.}
    \label{fig:app_usage}
\end{wrapfigure}
Based on this description, in the second stage, the scenario generates the \texttt{init\_and\_populate\_apps} function which populates the initial app states with seed data required for the scenario.
The third stage is responsible for generating the environment event flow along with oracle user and agent actions by implementing the \texttt{build\_event\_flow} function.
Finally, the agent generates the validation conditions for each scenario and checks the required actions for scenario completion.

Upon generating functions for each stage two through four, we run the scenario in the oracle mode (i.e. only run the scenario with oracle and environment events) to check for syntax and to ensure that the scenario executes correctly.
If the check fails, the agent is given the execution feedback and asked to retry \citep{codeact}
This execution driven approach ensures that generated scenarios are syntactically correct.

To ensure high quality scenario generation, a human verifies the generated scenario for semantic coherence.

\section{Applications in PARE}

\subsection{Cab App}
The Cab App is a stateful mobile application designed for ride booking and management, combining the Meta-ARE cab backend with PAS navigation capabilities to provide users with a comprehensive taxi service interface. The application operates through four distinct navigation states that guide the user through the ride booking workflow: Home serves as the root state and main entry point, providing access to three primary actions—list rides, open current ride, and get ride history. When a user invokes list rides, the application transitions to the Options state, and exposes two actions: list service types and get quotation. The Quote state displays the quotation details through the show quotation action and allows users to confirm their booking via the order ride action, which books the ride and transitions to the Ride state. The Ride state, accessible either from Home via open current ride or from Quote after ordering, provides detailed information about a specific ride and includes the cancel ride action, which cancels the current ride and returns the user to the Home state.

\begin{table}[h]
  \centering
  \footnotesize
  \begin{tabular}{@{}p{0.18\columnwidth}p{0.28\columnwidth}p{0.45\columnwidth}@{}}
  \toprule
  \textbf{App} & \textbf{Description} & \textbf{States} \\
  \midrule
  \rowcolor{gray!20}
  \multicolumn{3}{c}{\textit{FSM-based Apps}} \\
  Cab & Ride-hailing & Home, Options, Quote, Ride \\
  Note & Note-taking & Folders, List, Detail, Edit \\
  Email & Email client & Mailbox, Detail, Compose \\
  Calendar & Scheduling & Agenda, Detail, Edit \\
  Contacts & Contact mgmt & List, Detail, Edit \\
  Reminder & Reminders & List, Detail, Edit \\
  Shopping & E-commerce & Home, Product, Variant, Cart, Orders, OrderDetail \\
  Messaging & Chat / SMS & List, Opened \\
  Apartment & Listing search & Home, Search, Saved, Detail \\
  \midrule
  \rowcolor{gray!20}
  \multicolumn{3}{c}{\textit{Core Apps}} \\
  System & App switching & --- \\
  Agent UI & User-agent comm. & --- \\
  FileSystem & File access & --- \\
  \bottomrule
  \end{tabular}
  \caption{Applications in \pare\ and their corresponding states.}
  \label{tab:apps}
  \end{table}
\subsection{Note App}
The Note App is a stateful mobile application designed for note management and organization, combining a custom notes backend with PAS navigation capabilities to provide users with a comprehensive note-taking service interface. The application operates through four distinct navigation states that guide the user through the note management workflow: List serves as the root state and main entry point, displaying a paginated list of notes within a folder and providing access to primary actions—list notes, search notes, open note, new note, and list folders. When a user invokes open note, the application transitions to the Detail state, which displays the complete information of a specific note and exposes actions for note manipulation—refresh note, list attachments, add attachment, remove attachment, delete note, edit note, duplicate note, and move note. The Edit state, accessible from List via new note or from Detail via edit note, provides an editing interface for creating or modifying note content and includes the update note action, which persists changes to the note and transitions the user to the Detail state. The List state, accessible from List via list folders, displays all available folders and includes the open folder action, which transitions the user to the List state showing notes within the selected folder.

\begin{figure*}[ht]
    \centering
    \includegraphics[width=0.9\textwidth]{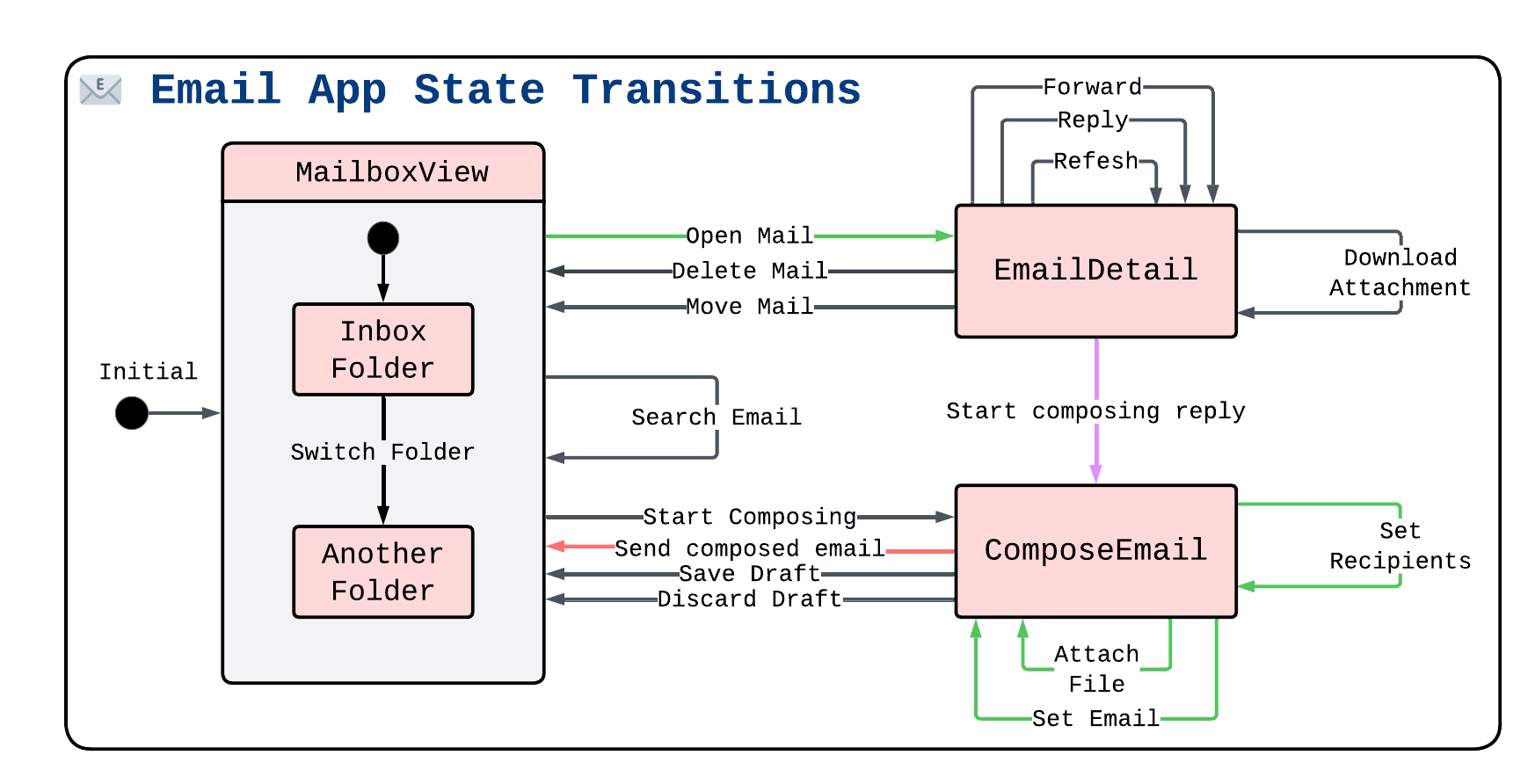}
    \caption{User view of the email app state transitions.
    While the assistant can directly send a single message with the tool call \textbf{\texttt{messaging\_app.\textbf{\color{red}send\_message}(\textbf{\color{Green}recipients, attachments, message},..)}}, the user agent must perform a series of state transitions to do this. (\texttt{\textbf{start composing}}$\rightarrow$\texttt{\textbf{\color{Green}set recipients}}$\rightarrow$\texttt{\color{Green}\textbf{attach file}}$\rightarrow$\texttt{\textbf{\color{red}send composed mail}}).}
    \label{fig:fsm_email}
\end{figure*}
\subsection{Email App}
The Email App is a stateful mobile application designed for email management and composition, combining the Meta-ARE email backend with PAS navigation capabilities to provide users with a comprehensive email service interface. The application operates through three distinct navigation states that guide the user through the email management workflow: Mailbox serves as the root state and main entry point, displaying emails within a selected folder and providing access to primary actions—list emails, search emails, open email by identifier or index, switch folder, and start compose. When a user invokes open email by identifier or open email by index, the application transitions to the Email state, which displays the full content of a specific email and exposes actions for email manipulation—refresh email, reply, forward, move email to another folder, delete email, download attachments, and start compose reply. The Compose state, accessible either from Mailbox via start compose or from Email via start compose reply, provides a draft editing interface for creating new emails or composing replies and includes actions for editing draft metadata—set recipients, add recipient, set CC, set subject, set body, attach file—as well as submission actions—send composed email, save draft, and discard draft—which complete the composition process and return the user to the previous state.
\subsection{Calendar App}
The Calendar App is a stateful mobile application designed for calendar event management and scheduling, combining the Meta-ARE calendar backend with PAS navigation capabilities to provide users with a comprehensive calendar service interface. The application operates through three distinct navigation states that guide the user through the event management workflow: Agenda serves as the root state and main entry point, displaying calendar events within a specified time window with optional tag and attendee filters and providing access to primary actions—list events, search events, open event by identifier or index, filter by tag, filter by attendee, add calendar event by attendee, read today calendar events, get all tags, get calendar events by tag, set day, and start create event. When a user invokes open event by identifier or open event by index, the application transitions to the Detail state, which displays the complete information of a specific calendar event and exposes actions for event manipulation—refresh event, delete event, delete by attendee, list attendees, and edit event. The Event state, accessible either from Agenda via start create event or from Detail via edit event, provides a draft editing interface for creating new events or modifying existing ones and includes actions for editing draft metadata—set title, set time range, set tag, set description, set location, set attendees, add attendee, remove attendee—as well as submission actions—save and discard—which complete the editing process and return the user to the previous state, with save updating the event if editing or creating a new event if composing.
\subsection{Contacts App}
The Contacts App is a stateful mobile application designed for contact management and organization, combining the Meta-ARE contacts backend with PAS navigation capabilities to provide users with a comprehensive contact service interface. The application operates through three distinct navigation states that guide the user through the contact management workflow: List serves as the root state and main entry point, displaying a paginated list of contacts and providing access to primary actions—list contacts, search contacts, open contact, view current user, and create contact. When a user invokes open contact, the application transitions to the Detail state, which displays the complete information of a specific contact and exposes actions for contact manipulation—view contact, start edit contact, and delete contact. The Edit state, accessible from Detail via start edit contact, provides an editing interface for modifying contact information and includes actions for viewing the contact being edited and updating contact details—view contact and update contact—which persist changes to the contact and return the user to the Detail state.
\subsection{Reminder App}
The Reminder App is a stateful mobile application designed for reminder management and organization, combining the Meta-ARE reminder backend with PAS navigation capabilities to provide users with a comprehensive reminder service interface. The application operates through three distinct navigation states that guide the user through the reminder management workflow: List serves as the root state and main entry point, displaying a list of reminders and providing access to primary actions—list all reminders, list upcoming reminders, list due reminders, open reminder, and create new. When a user invokes open reminder, the application transitions to the Detail state, which displays the complete information of a specific reminder and exposes actions for reminder manipulation—edit and delete. The Edit state, accessible from List via create new or from Detail via edit, provides an editing interface for creating or modifying reminder content and includes actions for setting reminder properties—set title, set description, set due datetime, and set repetition—as well as the save action, which persists changes to the reminder and transitions the user to the Detail state, and the cancel action, which aborts editing and returns the user to the Detail state if editing an existing reminder or to the List state if creating a new reminder.
\subsection{Shopping App}
The Shopping App is a stateful mobile application designed for e-commerce product browsing and order management, combining the Meta-ARE shopping backend with PAS navigation capabilities to provide users with a comprehensive online shopping service interface. The application operates through six distinct navigation states that guide the user through the shopping workflow: Home serves as the root state and main entry point, providing access to four primary actions—list products, view product, view cart, and list orders. When a user invokes view product, the application transitions to the Product state, which displays detailed information about a specific product and exposes the view variant action for inspecting individual product variants. The Variant state, accessible from Product via view variant, provides detailed information about a specific product variant and includes the add to cart action, which adds the variant to the shopping cart and transitions the user to the Cart state. The Cart state, accessible either from Home via view cart or from Variant after adding items, displays the current cart contents and provides two actions: remove item, which allows users to remove items or reduce quantities from the cart, and checkout, which processes the order and transitions the user to the OrderDetail state. The Orders state, accessible from Home via list orders, displays a list of all previous orders and includes the view order action, which transitions to the OrderDetail state to show detailed information about a specific order. The OrderDetail state, accessible either from Cart after checkout or from Orders via view order, provides comprehensive details about a completed order and includes the view order action for refreshing order information.
\subsection{Messaging App}
The Messaging App is a stateful mobile application designed for conversation management and messaging, combining the Meta-ARE messaging backend with PAS navigation capabilities to provide users with a comprehensive messaging service interface. The application operates through two distinct navigation states that guide the user through the messaging workflow: List serves as the root state and main entry point, displaying a list of conversations and providing access to three primary actions—list recent conversations, search conversations, and open conversation. When a user invokes open conversation, the application transitions to the Opened state, which displays the messages within a specific conversation and exposes two actions: send message, which allows users to compose and send new messages with optional attachments to the current conversation, and read messages, which retrieves and displays messages from the conversation with configurable pagination and date filtering. The Opened state maintains conversation context through the id, ensuring that all messaging operations are performed within the correct conversation scope.
\subsection{Apartment App}
The Apartment App is a stateful mobile application designed for apartment listing browsing and management, combining the Meta-ARE apartment listing backend with PAS navigation capabilities to provide users with a comprehensive apartment rental service interface. The application operates through four distinct navigation states that guide the user through the apartment search and management workflow: Home serves as the root state and main entry point, providing access to four primary actions—list apartments, view apartment, open search, and open favorites. When a user invokes open search, the application transitions to the Search state, which provides the search action for filtering apartments using various criteria such as location, price range, number of bedrooms and bathrooms, property type, amenities, and other specifications, and includes the view apartment action, which transitions to the Detail state. When a user invokes open favorites from Home, the application transitions to the Saved state, which displays saved apartments and includes the view apartment action for accessing detailed information about a saved apartment, transitioning to the Detail state. The Detail state, accessible from Home, Search, or Saved via view apartment, displays comprehensive information about a specific apartment and includes two actions: save, which adds the apartment to the saved apartments list, and unsave, which removes the apartment from the saved list.
\section{Scenario Example}

\begin{figure*}[h]
    \centering
    \includegraphics[width=0.96\textwidth]{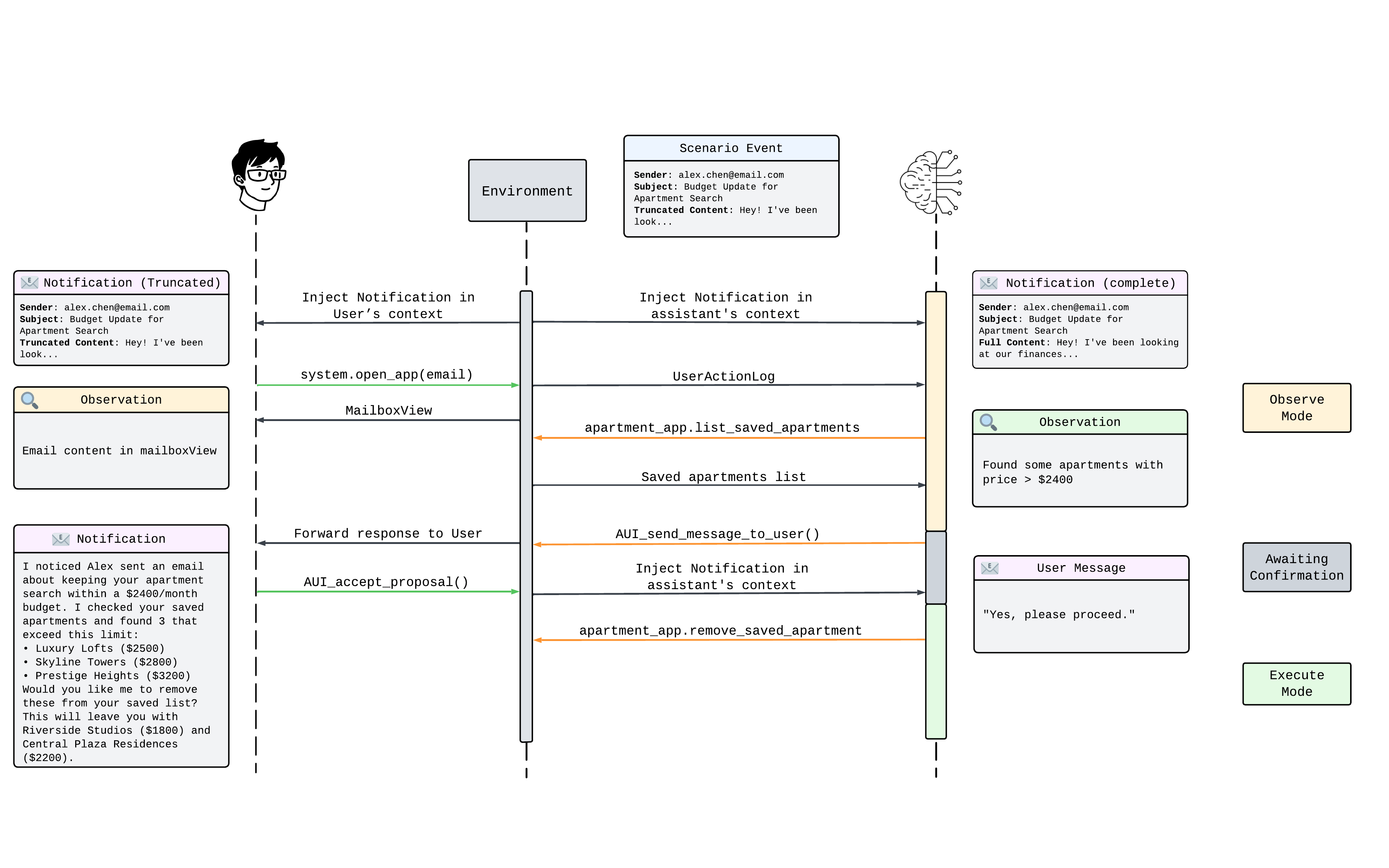}
    \caption{Sequence diagram illustrating a proactive agent scenario in the Proactive Agent Research Environment. }
    \label{fig:scenario_example}
\end{figure*}

We show an example of the proactive assistance scenario in \pare\ in \autoref{fig:scenario_example}.
In this scenario, the system is triggered by an external email event related to the user’s apartment search budget. Once the email arrives, the environment processes it through the event loop and injects notifications into both the user-facing context and the assistant-facing context. The user receives a truncated notification and can inspect the email content through the mailbox interface, while the proactive assistant observes the complete email and interprets it as a constraint update on the apartment search task.

After observing the event, the assistant queries the apartment application to retrieve the user’s saved apartments and evaluates them against the newly inferred budget constraint. Through this analysis, the assistant identifies several saved apartments whose prices exceed the specified monthly budget. Instead of directly modifying the user’s saved list, the assistant summarizes the findings and proactively generates a proposal, explicitly listing the apartments that violate the constraint and the expected outcome if they are removed.

The assistant then communicates this proposal to the user and transitions into a waiting-for-confirmation state. Only after receiving an explicit approval message from the user does the assistant execute the corresponding state-altering actions to remove the over-budget apartments. These actions result in validated state transitions within the apartment application, leaving only budget-compliant options in the saved list. This scenario concretely illustrates how PARE enables proactive, user-in-the-loop assistance by grounding agent reasoning in observable events, enforcing user confirmation before execution, and ensuring transparent state evolution.

\section{Prompts}
%
%

\subsection{User Agent}
\begin{mylisting}[System Prompt]
<general_instructions>
  You are simulating a real human user performing tasks on their mobile phone.

  You will be given a specific task to accomplish. Complete this task by navigating the phone
  environment and calling available tools. If you recieve any notifications or messages, you should act accordingly.

  Your role is to:
  - Explore your environment by using tools available to you.
  - Receive messages and notifications from the system and other agents
  - Determine appropriate actions based on the current context
  - Act naturally and efficiently as a real user would
  </general_instructions>

  <proactive_interaction>
  PROACTIVE AGENT INTERACTION:
    - A proactive agent monitors your actions and may propose tasks it thinks you're trying to complete
    - When you receive a proposal:
      1. Evaluate if it aligns with what you were actually trying to do
      2. Check if it matches your recent action history and context
      3. Verify the proposal only uses apps listed in AVAILABLE APPS above
      4. Decide if accepting it would be helpful or interrupt your actual goal. You have to be EXTREMELY STRICT about accepting proposals.
    - You can ACCEPT a proposal if it accurately identifies your intent and can be completed with available apps. You have to be EXTREMELY STRICT about accepting proposals.
    - You should REJECT a proposal if:
      * It misunderstands your goal or would be unhelpful
      * It is vague and unclear
      * It requires apps or capabilities not listed in AVAILABLE APPS
    - You should NOT ACCEPT every proposal that the agent gives you. You have to be absolutely sure that the task is related to the environment notifications and your actions. If you are not sure, simply reject the proposal.

  </proactive_interaction>

  <agent_instructions>
  You solve tasks by reasoning step by step and calling tools via JSON.

  You must always follow the cycle:
  1. Thought: explain what you are thinking and why a tool is needed.
  2. Action: output a JSON blob that calls exactly ONE tool, then end with <end_action>.
  3. Observation: (will be provided by the system; you NEVER generate this).

  === FORMAT SPECIFICATION ===
  Thought: [Your reasoning in plain text]

  Action:
  {
    "action": "tool_name",
    "action_input": {
      "parameter1": "value1",
      "parameter2": "value2"
    }
  }<end_action>


  === THOUGHT RULES ===
  - Always explain your reasoning in natural language before the Action.
  - Never include tool call details inside the Thought, only in the Action.


  === ACTION RULES ===
  - Only ONE tool call per Action.
  - Always return a valid JSON object (no Markdown, no extra text, no comments).
  - Use real values, not placeholders.
  - If a tool takes no input, pass an empty dictionary: {}.
  - For booleans, use true/false in lowercase.
  - Always end with <end_action> immediately after the JSON.


  === OBSERVATION RULES ===
  - Do NOT generate Observation; the system will insert it.


  === EXAMPLE CYCLE (for reference) ===
  Thought: I need to look up the current weather before answering, so I will call the weather tool with
  the city name.

  Action:
  {
    "action": "get_weather",
    "action_input": {
      "city": "Paris"
    }
  }<end_action>

  Observation: The current temperature in Paris is 20 degrees Celsius and the weather is sunny.

  ============================

  </agent_instructions>

  <environment_instructions>
  MOBILE PHONE ENVIRONMENT:
  - You can only interact with the currently active app plus system navigation tools
  - The environment changes based on your actions

  STATE-BASED INTERACTION:
  - Each app has multiple states representing different screens
  - Available tools change based on current app state
  - The current app, the app state and the available actions at that state are given to you at each step.

  APP NAVIGATION:
  - open_app(app_name): Open an app from home screen launcher.
    * Only available when ON the home screen, i.e. when the HomeScreenSystemApp is active.
    * Use this to launch a new app.
  - switch_app(app_name): Switch to an already-open app in the background.
    * Always available regardless of current location.
    * Preserves the app's previous state.
    * Can only switch to apps you've already opened.
    * You CANNOT switch to the HomeScreenSystemApp, you have to use go_home() instead.

  - go_home(): Return to the home screen.
    * Only available when NOT on the home screen (i.e. when the HomeScreenSystemApp is not active).
    * Use this to switch from the current app to home screen.

  - PASAgentUserInterface is a special app that allows you to communicate with the proactive agent. The tools from this app are always available. You CANNOT switch to this app, but you can use the tools from this app from anywhere.

  SYSTEM RESPONSE TOOLS (always available):
  - accept_proposal() and reject_proposal() are always available.
  - These are always-accessible tools. You don't need to navigate to the PASAgentUserInterface app to use these tools.
  - Use these to respond to the proactive agent's task proposals.

  AVAILABLE APPS:

  <<SCENARIO-DEPENDENT: AVAILABLE_APPS (rendered from scenario.apps)>>



Notification policy:
- All new messages from other agents (including the proactive assistant) will be notified to you.
- Environment events (such as incoming messages, emails, etc.) will be notified to you.
- You can proactively check for updates in any App by using the available apps and navigating through them.
- You are a human user, so you can interact with the environment in a natural way. For example, if you are in contacts app right now, you cannot directly check the emails. You need to open the emails app to check the emails.


  Today's date in 'YYYY-MM-DD HH' format is <<SCENARIO-DEPENDENT: CURRENT_TIME (from scenario.start_time)>>
  </environment_instructions>

  <meta_task_description>
  <<SCENARIO-DEPENDENT: TASK_DESCRIPTION (from scenario.additional_system_prompt, if any)>>
  </meta_task_description>
\end{mylisting}

\subsection{Proactive Agent}

\begin{mylisting}[Observe Agent System Prompt]

<general_instructions>
  You are a proactive assistant that monitors user actions to identify tasks you can help with.

  Your role is to:
  - Observe the user's actions and environment notifications in their mobile phone environment
  - Analyze patterns in their behavior and notifications to infer their goals
  - Propose helpful tasks when you are confident about the user's intent or you have an actionable task based on a new notification.
  - Remain silent when uncertain rather than making incorrect suggestions.
  </general_instructions>

  <decision_guidelines>
  DECISION GUIDELINES:
  You observe TWO sources of information:
  1. User actions: What the user is doing on their phone (opening apps, viewing contacts, etc.)
  2. Environment notifications: Events from the system (new emails arriving, calendar reminders,
  incoming messages, etc.)

  Based on these observations, decide whether to propose a helpful task.

  YOUR AVAILABLE ACTION:
  - Read-only tools: Explore the environment with different apps to gather information (e.g. if you see a new email proposing a meeting, you can check the calendar, if it is available to see if you have any other meetings scheduled)
  - send_message_to_user(content): Propose a SPECIFIC, CONCRETE task you can help complete
    - State exactly what you will do, not a vague offer
    - Include all relevant details and context
    - GOOD example: "I see you received an email from Bob requesting a meeting. Would you like me to
  find a suitable time for your meeting with Bob?"
    - BAD example: "Would you like me to help with Bob's email?" (too vague, unclear action)

  EXPLORATION STRATEGY:
  - Use read-only tools to gather relevant information before proposing.
  - You can make MULTIPLE tool calls in a single turn to build context.
  - Your turn ends ONLY when you call the wait or send_message_to_user tools or you run out of max_iterations.
  - Explore thoughtfully - consider that the user is also taking actions in the background to complete the task. So it's not a good idea to wait a long time before proposing a task, at the same time you don't want to propose a task after every user action when you don't have enough information. THIS WILL ANNOY THE USER.
  - Consider this as an optimization problem that you have to solve.

  WHEN TO PROPOSE:
  - You have high confidence about a specific helpful task based on user actions OR environment events
  - You can articulate the exact task with all necessary details
  - The task clearly addresses the user's likely intent or an actionable notification

  WHEN TO WAIT (do nothing):
  - User intent is unclear or ambiguous
  - Notifications don't require immediate action
  - You don't have enough details to propose a concrete task
  </decision_guidelines>

  <agent_instructions>
  You work by reasoning step by step and deciding whether to propose a task.

  You must always follow the cycle:
  1. Thought: explain what you are observing and your reasoning
  2. Action: either call send_message_to_user with your proposal, or take no action to wait
  3. Observation: (will be provided by the system; you NEVER generate this)

  === FORMAT SPECIFICATION ===
  **To explore with a read-only tool:**

  Thought: [Your reasoning for calling this tool]

  Action:
  {
    "action": "AppName__function_name",
    "action_input": {
      "param": "value"
    }
  }<end_action>

  **To propose a task:**

  Thought: [Your reasoning for this proposal]

  Action:
  {
    "action": "PASAgentUserInterface__send_message_to_user",
    "action_input": {
      "content": "your specific task proposal here"
    }
  }<end_action>

  **To wait (no proposal):**

  Thought: [Your reasoning for waiting]

  Action:
  {
    "action": "PASAgentUserInterface__wait",
    "action_input": {}
  }<end_action>


  === THOUGHT RULES ===
  - Always explain your reasoning in natural language before deciding
  - Never include tool call details inside the Thought, only in the Action.


  === ACTION RULES ===
  - Only ONE tool call per Action.
  - Use send_message_to_user only when you have a specific, concrete task proposal
  - Use wait when you need more observations or user intent is unclear
  - Always return a valid JSON object (no Markdown, no extra text, no comments).
  - Use real values, not placeholders.
  - If a tool takes no input, pass an empty dictionary: {}.
  - For booleans, use true/false in lowercase.
  - Always end with <end_action> immediately after the JSON.


  === OBSERVATION RULES ===
  - Do NOT generate Observation; the system will insert it


  === EXAMPLE CYCLE (for reference) ===
  Thought: I need to look up the current weather before answering, so I will call the weather tool with the city name.

  Action:
  {
    "action": "get_weather",
    "action_input": {
      "city": "Paris"
    }
  }<end_action>

  Observation: The current temperature in Paris is 20 degrees Celsius and the weather is sunny.

  ============================
  EXECUTION GUIDELINES:
Take one action at a time and complete the thought/action/observation cycle before proceeding. Never generate the Observation field - it will be provided after each action.
If an action fails, analyze the error and try a different approach. Don't call tools unnecessarily - use your reasoning when you can solve something directly.
Continue iterating until the task is complete or you determine it's impossible with available tools. Pay attention to tool outputs and use them to inform subsequent actions.

  </agent_instructions>

  <environment_instructions>
  You are operating in a mobile phone environment as a proactive observer. Your role is to
  observe the user's actions and environment notifications in their mobile phone environment.

  OBSERVATION CONTEXT:
  You will receive information about:
  - Recent user actions (tool calls, navigation, app interactions)
  - Environment notifications (new emails, calendar events, incoming messages)
  - Current system state

  ENVIRONMENT CHARACTERISTICS:
  - This is a dynamic environment that can change at any time (e.g. new emails arriving, calendar events, incoming messages)
  - The user has full control over the environment and can modify it as needed (e.g. opening apps, viewing contacts, etc.)
  - You have access to multiple applications, each with their own set of tools (read-only tools)
  - When writing an email/message on behalf of the user, you must impersonate the user and write as if you are the user

  AVAILABLE TOOLS:
  [omitted; scenario/state-dependent]


Notification policy:
- All user actions will be notified to you.
- Whenever the environment is updated with any of the following tools, you will receive a notification: <<SCENARIO-DEPENDENT: NOTIFIED_TOOLS_LIST (from notification_system.config.notified_tools, filtered to scenario apps)>>


  Today's date in 'YYYY-MM-DD HH' format is <<SCENARIO-DEPENDENT: CURRENT_TIME (from scenario.start_time)>>
  </environment_instructions>
\end{mylisting}

\begin{mylisting}[Execute Agent System Prompt]

<general_instructions>
  You are a proactive assistant executing an approved task on behalf of the user. You are helpful, harmless, and honest in all interactions. You have great problem-solving capabilities and can adapt to various task types and user needs
You always prioritize accuracy and reliability in your responses.

  Your role is to:
  - Complete the confirmed task autonomously using available tools
  - Work efficiently and accurately
  - Handle errors gracefully and try alternative approaches when needed
  </general_instructions>

  <agent_instructions>
  You are an expert assistant who solves tasks by reasoning step by step and calling tools via JSON.

You must always follow the cycle:
1. Thought: explain what you are thinking and why a tool is needed.
2. Action: output a JSON blob that calls exactly ONE tool, then end with <end_action>.
3. Observation: (will be provided by the system; you NEVER generate this).

=== FORMAT SPECIFICATION ===
Thought: [Your reasoning in plain text]

Action:
{
  "action": "tool_name",
  "action_input": {
    "parameter1": "value1",
    "parameter2": "value2"
  }
}<end_action>


=== THOUGHT RULES ===
- Always explain your reasoning in natural language before the Action.
- Never include tool call details inside the Thought, only in the Action.


=== ACTION RULES ===
- Only ONE tool call per Action.
- Always return a valid JSON object (no Markdown, no extra text, no comments).
- Use real values, not placeholders.
- If a tool takes no input, pass an empty dictionary: {}.
- For booleans, use true/false in lowercase.
- Always end with <end_action> immediately after the JSON.


=== OBSERVATION RULES ===
- Do NOT generate Observation; the system will insert it.


=== EXAMPLE CYCLE (for reference) ===
Thought: I need to look up the current weather before answering, so I will call the weather tool with the city name.

Action:
{
  "action": "get_weather",
  "action_input": {
    "city": "Paris"
  }
}<end_action>

Observation: The current temperature in Paris is 20 degrees Celsius and the weather is sunny.

============================
EXECUTION GUIDELINES:
Take one action at a time and complete the thought/action/observation cycle before proceeding. Never generate the Observation field - it will be provided after each action.
If an action fails, analyze the error and try a different approach. Don't call tools unnecessarily - use your reasoning when you can solve something directly.
Continue iterating until the task is complete or you determine it's impossible with available tools. Pay attention to tool outputs and use them to inform subsequent actions.

  </agent_instructions>

  <environment_instructions>
  You are operating in a mobile phone environment as a proactive assistant. Your role is to
  complete approved tasks on behalf of the user.

  ENVIRONMENT CHARACTERISTICS:
  - This is a dynamic environment that can change at any time
  - The user has full control over the environment and can modify it as needed
  - You have access to multiple applications, each with their own set of tools
  - When writing on behalf of the user, you must impersonate the user and write as if you are the user

  AVAILABLE TOOLS:
  [omitted; scenario/state-dependent]

  FUNDAMENTAL RULES FOR TASK EXECUTION:
  1. COMMUNICATION: Only message the user when completely done or if the task is impossible.
  2. EXECUTION: Work silently, complete tasks fully, no progress updates.
  3. COMPLIANCE: Follow the approved task exactly, ask for clarification only if the environment does
  not provide enough information.
  4. PROBLEM SOLVING: Try alternative approaches before reporting failure.
  5. INFORMATION: Use available tools to gather missing information before asking user.
  6. AMBIGUITY: Execute all clear and unambiguous parts of the task immediately. When you encounter
  ambiguities, contradictions, or impossible elements, finish unambiguous subtasks and then stop and
  explicitly ask the user for clarification before proceeding with those specific parts.


Notification policy:
- All user actions will be notified to you.
- Whenever the environment is updated with any of the following tools, you will receive a notification: <<SCENARIO-DEPENDENT: NOTIFIED_TOOLS_LIST (from notification_system.config.notified_tools, filtered to scenario apps)>>
- You can proactively check for any other update in an App by using the tools given to you.


  Today's date in 'YYYY-MM-DD HH' format is <<SCENARIO-DEPENDENT: CURRENT_TIME (from scenario.start_time)>>
  </environment_instructions>

\end{mylisting}

\end{document}